%% file: main.tex
\documentclass[10pt, a4paper]{article}
\usepackage[a4paper,left=0.8in,top=0.8in,right=0.8in,bottom=0.8in]{geometry}

\usepackage[utf8]{inputenc} 
\usepackage[T1]{fontenc}    
\usepackage{hyperref}       
\usepackage{url}            
\usepackage{nicefrac}       
\usepackage{microtype}      
\usepackage{subfigure}
\usepackage{booktabs} 

\usepackage{newcent} 
\usepackage{amssymb,amsmath,amsfonts}
\usepackage{graphicx}   
\usepackage{verbatim}   
\usepackage{color}      
\usepackage{stmaryrd}
\usepackage{bbm}
\usepackage{multirow}
\usepackage{pifont}
\usepackage{textcomp}
\usepackage{dsfont}

\usepackage[normalem]{ulem}
\useunder{\uline}{\ul}{}
\usepackage{amsmath}
\usepackage{amsfonts}
\usepackage{amsthm}
\usepackage{graphicx}
\usepackage{tikz,mathtools}
\usepackage{caption}
\usepackage{indentfirst}
\usepackage{algorithm}
\usepackage{algorithmic}

\usepackage{hyperref}       
\hypersetup{
	colorlinks  = true,    
	urlcolor     = blue,   
	linkcolor   = blue,    
	citecolor    = blue       
}

\usepackage[capitalize]{cleveref}

\newtheorem{theorem}{Theorem}
\newtheorem{definition}{Definition}
\newtheorem{lemma}{Lemma}

\newtheorem{remark}{Remark}
\newtheorem{assumption}{Assumption}

\newenvironment{myproof}[1]{
	
		\noindent
		\hrulefill
		
		\vspace{1mm}
		\noindent\textbf{Proof #1: }				
		
		\vspace{-2mm}
		\noindent
		\hrulefill
		
	}
	{$\hfill\square$
		
		\vspace{-2mm}
		\noindent
		\hrulefill
		
	\bigskip
}

\input{definitions.tex}

\usepackage[colorinlistoftodos, textwidth=4cm, shadow]{todonotes}

\newcommand{\vertiii}[1]{{\left\vert\kern-0.25ex\left\vert\kern-0.25ex\left\vert #1
    \right\vert\kern-0.25ex\right\vert\kern-0.25ex\right\vert}}

\newcommand{\key}[1]{\emph{#1}}

\newcommand{\Sam}{\cS}
\newcommand{\CI}{\textsf{CI}}
\newcommand{\SCI}{\textsf{SCI}}

\title{\textsc{How to Shrink Confidence Sets for Many Equivalent Discrete Distributions?}}

\author{Odalric-Ambrym Maillard\footnote{Scool Team, Centre Inria de l'Université de Lille (email: odalric.maillard@inria.fr)} ~~~~ Mohammad Sadegh Talebi\footnote{Department of Computer Science, University of Copenhagen (email: m.shahi@di.ku.dk)}}

\begin{document}

\maketitle

\begin{abstract}
	We consider the situation when a learner faces a set of unknown discrete distributions $(p_k)_{k\in \cK}$ defined over a common alphabet $\cX$, and can build for each distribution $p_k$ an individual high-probability confidence set thanks to $n_k$ observations sampled from $p_k$. The set $(p_k)_{k\in \cK}$ is structured: each distribution $p_k$ is obtained from the same common, but unknown, distribution q via applying an unknown permutation to $\cX$. We call this \emph{permutation-equivalence}. The goal is to build refined confidence sets \emph{exploiting} this structural property. Like other popular notions of structure (Lipschitz smoothness, Linearity, etc.) permutation-equivalence naturally appears in machine learning problems, and to benefit from its potential gain calls for a specific approach. We present a strategy to effectively exploit permutation-equivalence, and provide a finite-time high-probability bound on the size of the refined confidence sets output by the strategy. Since a refinement is not possible for too few observations in general, under mild technical assumptions, our finite-time analysis establish when the number of observations $(n_k)_{k\in \cK}$ are large enough so that the output confidence sets improve over initial individual sets. We carefully characterize this event and the corresponding improvement. Further, our result implies that the size of confidence sets shrink at asymptotic rates of $O(1/\sqrt{\sum_{k\in \cK} n_k})$ and $O(1/\max_{k\in K} n_{k})$, respectively for elements inside and outside the support of q, when the size of each individual confidence set shrinks at respective rates of $O(1/\sqrt{n_k})$ and $O(1/n_k)$. We illustrate the practical benefit of exploiting permutation equivalence on a reinforcement learning task.
\end{abstract}

\section{Introduction}

Like Lipschitz smoothness, linearity or sub-modularity, leveraging a \emph{structural property} of 
a set of unknown distributions, that can be known only by sampling, is generally the key to better statistical efficiency, hence improved learning guarantees. 
In this paper, we consider a learning task involving a set of unknown distributions 
$(p_k)_{k\in\cK}$ over a discrete alphabet $\cX\subset\Nat$ that are known to satisfy a structural property called \key{permutation-equivalence}.
Intuitively, this means all the distributions are actually the same up to a shuffling of the entries.
Formally, permutation-equivalence means that there exists a common distribution $q$ over $\cX$ such that each distribution $p_k, k\in\cK$ is obtained from $q$ after applying some permutation $\sigma_k$ of its entries (See Definition~\ref{def:perm_equivalence}).

Permutation-equivalence can be spotted in several situations.
For instance in a decentralized learning task where $\cK$ is a set of learners, different learners may number the observation space differently. Hence every process $q$ on the observation space will be seen as a different $p_k$ by learner $k$.
Permutation-equivalence also naturally appears in reinforcement learning. Indeed in several environments such as RiverSwim and grid-world Markov Decision Processes (2-room, 4-room, frozen-lake, etc.)\footnote{These environments are typically used as benchmarks in the reinforcement learning literature.}, probability transitions from two different state-action pairs are usually not arbitrarily different: In grid-worlds, the set of next-state transitions $(p(\cdot|s,a))_{s\in \cS_0,a\in \cA}$, where $\cS_0$ denotes all states with no neighboring wall, typically exhibits permutation-equivalence. This has been  considered in \cite{asadi2019model,lyu2023scaling}. Likewise,
a windy grid-world \cite{sutton1998reinforcement}, sailboat \cite{epshteyn2008active}, or RiverSail environment (see Appendix~\ref{app:riversail}) in which navigation is similar in each state or river channel except for the presence of a wind of unknown but constant direction also exhibit permutation-equivalence structure. Naturally, in practice a given task may present further structural properties beyond equivalence; we focus in this paper on the benefit that exploiting permutation-equivalence only can bring to the learner.

Exploiting permutation-equivalence, like any other structural assumption, is especially beneficial in when acquiring data from distributions is \emph{costly}. In the context of statistical estimation
where a learner has only access to $n_k$ samples from distribution $p_k$, for each $k\in\cK$,
by exploiting the structure we mean to build, using all samples, tighter confidence sets around each $p_k$ than 
the initial ``individual" confidence set using only samples from $p_k$ (that disregards structure).
The task is non-trivial when both $q$ and the underlying permutations connecting each $p_k$ to $q$ are \emph{unknown}. For instance identifying the structure  involves searching over a combinatorial space.
 This creates a trade-off between preserving statistical efficiency and providing a computational efficiency relaxation of an NP-hard problem.
 Also, it is intuitive that when initial confidence sets are too large (the $n_k$ are to small), it is hopeless to obtain any statistically valid tightening, while when they are all very sharp, a maximal tightening is reachable. We intend to precisely capture these transitory regimes in a problem-dependent way.

We answer this statistical versus computational trade-off question positively: We introduce a strategy of (low)-polynomial complexity in $|\cK|$ and $|\cX|$ that still achieves a non-trivial sample-efficiency speed-up. More precisely, we provide a finite-time high-probability bound on the size of the refined confidence sets output by the strategy, which enables to carefully establish when the number of observations $(n_k)_{k\in \cK}$ are large enough so that the output confidence sets improve over initial individual sets, and to quantify the corresponding tightening.  In particular, our result implies that the confidence sets exploiting equivalence shrink at asymptotic rate of $O(1/\sqrt{\sum_{k\in \cK} n_k})$ and $O(1/\max_{k\in K} n_{k})$, respectively for elements inside and outside the support of $q$, when the size of each individual confidence set shrinks at respective rates of $O(1/\sqrt{n_k})$ and $O(1/n_k)$. This improvement may be significant for a large set $\cK$.

\paragraph{Related work: Permutation-invariance and learning permutation.}
While estimation of discrete distribution has been intensively studied across many fields (\cite{cover2012elements}, \cite{weissman2003inequalities}, \cite{berend2013concentration}, \cite{mardia2019concentration}), the existing literature on estimation of distributions \emph{under some notion of equivalence}, up to our knowledge, is scarce. 
A rich literature exists on exploiting or enforcing the notion of \emph{permutation-invariance}, that is when the probability mass remains unchanged upon applying any permutation on the instances.
In \cite{hemerik2018false,hemerik2018permutation}, the authors consider the construction of  confidence sets in the context of \emph{multiple  testing}. The task is to construct \emph{simultaneous} confidence sets for the false discovery proportion, such that the joint distribution of $p$-values of the part of the data corresponding to the null hypotheses should be invariant under permutations. Similarly to the current work,  permutations are used in these works to model the dependency structure in the distributions generating the data and build confidence sets.
However, \emph{permutation-invariance} differs from permutation-equivalence. 
In a different study, for classification, \cite{shivaswamy2006permutation} presents  permutation-invariant SVMs that enforce that the classifier is invariant to permutations of the elements of each input; see also \cite{wimmer2010agnostically}.
On the combinatorial side, there is a rich literature on \emph{learning permutation} in online learning (e.g., \cite{helmbold2009,yasutake2011online,ailon2014bandit}) and ranking (e.g., \cite{lebanon2002cranking,ailon_AISTATS2014,shah2016permutation}), whose focus is to identify the \emph{best permutation} among all possible ones. Other popular setups include optimal matching of known distributions (optimal transport, \cite{villani2008optimal}), or structured decision making when output decisions are permutations. 
In our setup, the task is however not to find the best permutation among $(\sigma_k)_{k\in \cK}$. Rather we want to identify permutations that, given the finite amount of samples at hand, allow us to \emph{combine samples} in a statistically efficient way. In other words, our primary goal is to build data-dependent sets with statistical guarantees. 
From the literature on bipartite graph and perfect matchings, such works as \cite{Fukuda94} may serve to build candidate permutations, although its complexity grows with the total number of perfect matchings.


\paragraph{Outline and contribution.} 
After briefly presenting permutation-equivalence (Definition \ref{def:perm_equivalence}) and intuitions, we present in Section~ \ref{sec:algo} an algorithm that, given a set of valid confidence sets of each considered distributions, identifies a set of admissible matchings between various distributions to group their corresponding samples. We illustrate on a simple example the potential benefit of the method in tightening confidence sets. 
In Section~\ref{sec:statbene}, we provide a finite time analysis 
showing precisely when and how tightening of confidence sets happen in a problem-dependent way. This is summarized in Theorems~\ref{thm:permutationequivalencegain} and \ref{thm:permutationequivalencegain2}. They imply for instance that the size of confidence sets output by the proposed algorithm roughly shrinks in $\cO(1/\sqrt{\sum_{k} n_k})$ and $\cO(1/\max_{k} n_{k})$, respectively for points inside and outside of the support, when the initial confidence sets size shrink at respective  rates of $\cO(1/\sqrt{n_k})$ and $\cO(1/n_k)$. This can be significant when $|\cK|$ is large. 
The results reported in Theorems~\ref{thm:permutationequivalencegain} and \ref{thm:permutationequivalencegain2} are valid for a broad class of distributions for which empirical confidence sets are available,
 and rely on a notion of \emph{surrogate} confidence set (see Definition \ref{def:surr_CI} and examples at the end of Section~\ref{sec:statbene}), which could be of independent interest. Last we illustrate the use of this approach on a simple RL task.

%

\section{Setup and Notations: Tightening Estimation using Equivalence}\label{sec:setup}

For a finite alphabet $\cX\subset\Nat$ we denote by $\cP(\cX)$ the set of probability distributions over $\cX$ and by $\bG_\cX$ the group of permutations over $\cX$. Each permutation $\sigma\in \bG_\cX$ acts as perfect matching (one-to-one mapping of $\cX$). We consider  a set $(p_k)_{k\in\cK}\subset\cP(\cX)$ of $K=|\cK|$ many distributions on $\cX$.
We assume they are all generated from the same common underlying distribution, after applying different permutation of $\cX$. More formally, we introduce the following definition:

\begin{definition}[Permutation-equivalent set]
	\label{def:perm_equivalence}
	Distributions $(p_k)_{k\in\cK}$ are said to be \emph{equivalent} under $\bG_\cX$ (\emph{\bf $\bG_\cX$-equivalent}), if there exists a common distribution $q$ such that each $p_k$ is obtained by applying one of the permutation from $\bG_\cX$ to $q$, namely,
	\beqa
	\label{eq:perm_equivalence}
	\exists q\in\cP(\cX),(\sigma_k)_{k\in\cK}\subset \bG_\cX :  \forall k\in\cK,  \,\,p_k  = q \circ \sigma_k\,.
	\eeqa
\end{definition}

We call  $q$ the \emph{canonical distribution} of $(p_k)_{k\in \cK}$, and $(q,(\sigma_k)_{k\in \cK})$ its \emph{canonical representation}.

\begin{remark}
Definition~\ref{def:perm_equivalence} naturally extends beyond finite $\cX$ and permutations, to any set $\cX$ with corresponding set $\cG_\cX$ of deformations (automorphisms). It also extends beyond $\cP(\cX)$ to functions. 
\end{remark}

\paragraph{Empirical estimates and confidence sets.}
The learner does not know the distributions $(p_k)_{k\in\cK}$. Instead, for each $k\in\cK$, a sample $(X_{k,i})_{i\leq n_k}$ of $n_k$ many i.i.d.~observations from $p_k$ is given, to form the empirical distribution $\hat p_k$. More formally, 
$$
\forall x\!\in\!\cX,\,\,
\hat p_k(x):=\hat p(\Sam_{k,x}) := \frac{1}{|\Sam_{k,x}|}\sum_{y\in \Sam_{k,x}}y\quad\text{where}\quad  \Sam_{k,x} = (\indic{X_{k,i}=x})_{i\leq n_k}\,.
$$
%
Further, the learner has access to a procedure $\CI$ to build confidence sets. For each given $\Sam_{k,x}$ and confidence level $\delta\!\in\!(0,1)$, it builds $c^{k}_x \!:=\!\CI(\Sam_{k,x},\delta)$ such that
$\forall k\!\in\!\cK, \Pr\Big(\exists x\!\in\!\cX,\, p_k(x) \!\notin\! c_x^k \Big)\!\leq\! \delta\,.$. Such sets are obtained only based on the sample $\Sam_{k,x}$, ignoring the $\bG$-equivalence structure.
Noting that $\Sam_{k,x}$ is a sample from a Bernoulli distribution with parameter $p_k(x)$, we consider confidence intervals written as follows:
\al{
	\label{eq:CI_generic}
	\CI({\Sam},\delta) =\Big\{ \lambda \in [0,1]: d(\hat p(\Sam),\lambda) \leq b(\lambda,\Sam,\delta)\Big\} \,,
}
where $d$ is some distance function, and $b$ is  a decreasing function of the sample size $|\Sam|$. For example, the confidence interval defined using Bernstein's concentration inequality for $[0,1]$-bounded observations (see \cite{boucheron}) uses $d(x,y) = |x-y|$ and
%
\als{
	b^{\texttt{Berns}}(\lambda,\Sam,\delta) = \sqrt{\frac{2\lambda(1\!-\!\lambda)\log\big(\tfrac{2|\cX|}{\delta}\big)}{|\Sam|}} + \frac{\log\big(\tfrac{2|\cX|}{\delta}\big)}{3|\Sam|} \, .
}
Equation~\eqref{eq:CI_generic} allows for greater flexibility, and examples are discussed in Appendix \ref{app:confexamples}.

\paragraph{Refined estimates and confidence intervals.}
The goal of the learner is to output novel confidence sets
$(\overline{\CI}_{k,x})_{k\in\cK,x\in \cX}$ for  $(p_k)_{k\in\cK}$ called the \key{refined confidence sets}, that exploit $\bG$-equivalence and make use of all samples $(\Sam_{k,x})_{k\in\cK,x\in \cX}$.
More precisely, these sets $(\overline{\CI}_{k,x})_{k\in\cK,x\in \cX}$ must satisfy: 
%
\beqan
\forall\delta\in(0,1),\,
\forall k\!\in\!\cK,\quad	\Pr\Big(\!\exists x\!\in\!\cX,\,\!p_k(x) \!\notin\! \overline{\CI}_{k,x}(\delta)\! \Big)\!\leq\! \delta.
\eeqan
%
Further, they must  
(i) not depend on any unknown quantity, and (ii) be of as small size as possible.

\vspace{-2mm}
\paragraph{Warming-up: $K=2$.} To conclude this section, we provide some insights in the simplified case of $K=2$ distributions, and discuss preliminary ideas to build refined confidence sets.

\begin{figure}[!hbtp]
	\centering
	\vspace{-3mm}
	\begin{minipage}{0.5\textwidth}
	\includegraphics[trim={0cm 0.1cm 0cm 3.5cm},clip,width=0.89\textwidth]{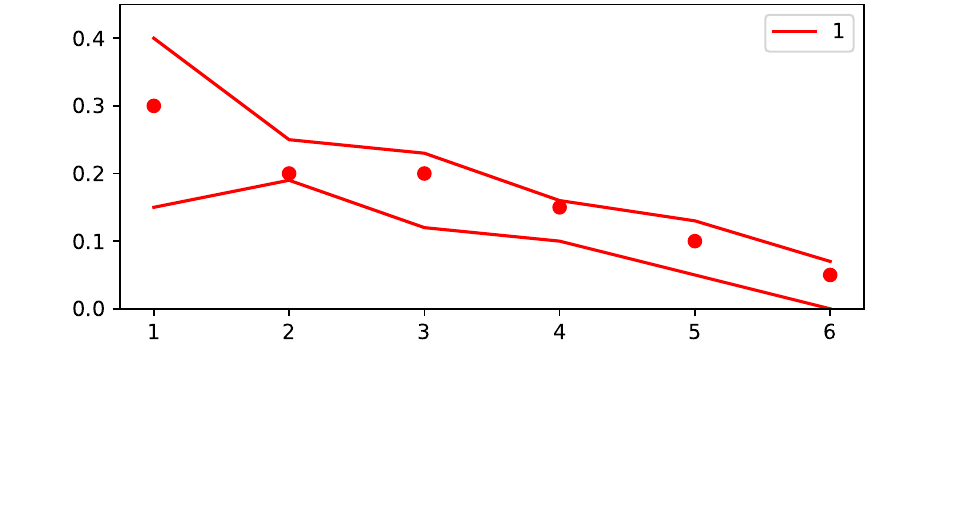}\\
	
	\vspace{-17mm}
	\includegraphics[trim={0cm 0.1cm 0cm 3.5cm},clip,width=0.89\textwidth]{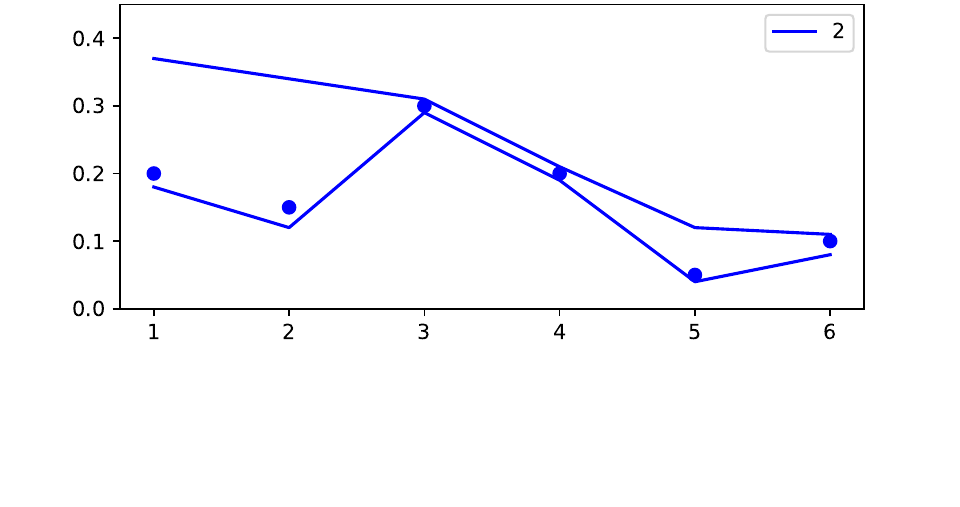}
	\end{minipage}\hfill
\begin{minipage}{0.5\textwidth}
		\includegraphics[trim={0cm 0.1cm 0cm 3.5cm},clip,width=0.89\textwidth]{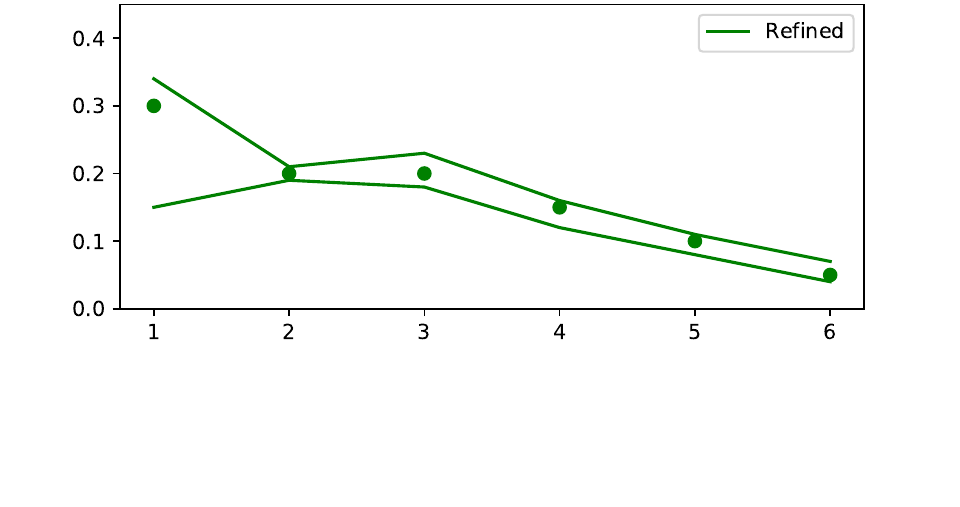}\\
		
	\vspace{-17mm}
	\includegraphics[trim={0cm 0.1cm 0cm 3.5cm},clip,width=0.89\textwidth]{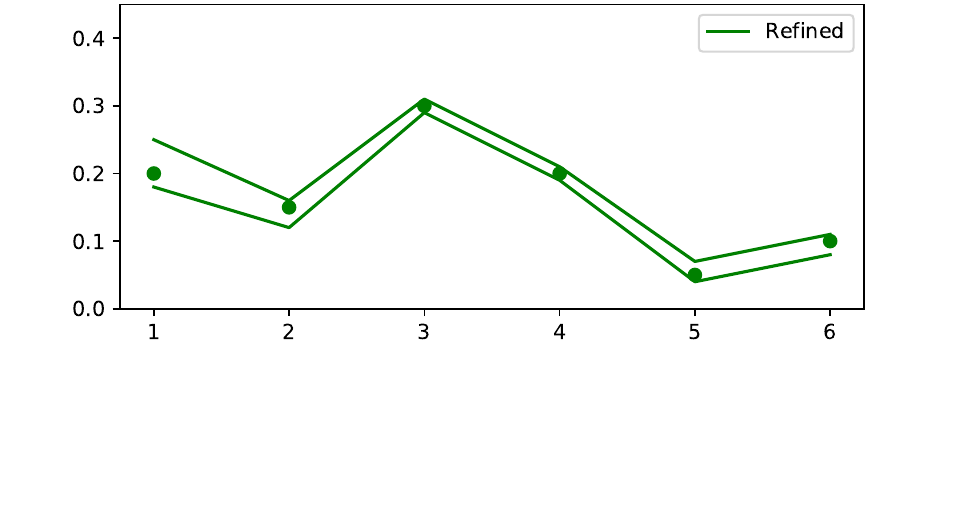}
	\end{minipage}

	\vspace{-11mm}
	\caption{Left: two equivalent distributions  (red, blue dots) and their Upper and Lower confidence bounds.
	Right: Refined confidence bounds exploiting equivalence.}
	\label{fig:twoprofiles}
\end{figure}

Figure~\ref{fig:twoprofiles} provides an example of two distributions (shown in blue and red), defined on the same alphabet of size 6. Their  true (unknown) values are depicted, together with the initial (not refined) confidence interval built from some samples.
Let ${\color{red}c_i}$ denote the confidence interval at point $x_i$ for the top (red) distribution,  and ${\color{blue}d_i}$ for the down distribution.
We observe that ${\color{blue}d_1}$  has non-empty intersection with ${\color{red}c_1},{\color{red}c_2},{\color{red}c_3}$. More generally, 
one can consider all possible intersections compatible with $\bG_\cX$ equivalence (see Figure~\ref{fig:mixprofiles}). This can used to build a bipartite graph with class blue and red, and edges showing non-empty intersections. From Figure~\ref{fig:mixprofiles}, ${\color{blue}d_3}$  can only match ${\color{red}c_1}$, from which we deduce that ${\color{blue}d_1,d_4}$ can only match ${\color{red}c_2,c_3}$. Hence,
due to the one-to-one assignment,  ${\color{blue}d_2}$ can only match ${\color{red}c_4,c_5}$. The same holds for ${\color{blue}d_6}$, which implies that ${\color{blue}d_5}$ is only compatible with ${\color{red}c_6}$. Hence, in this case, although the confidence intervals are not especially tight, it is possible to show that if $p_1=q$ and $p_2=q\circ \sigma_2$, then $\sigma_2$ can only be one of out of four possible permutations (out of 720 candidates). 
Further, we can easily tighten the confidence bounds based on the pruned sets of compatible matchings. For instance, from Figure~\ref{fig:mixprofiles}, the confidence interval on point $x_2$  becomes ${\color{blue}d_2}\cap{\color{red}c_4}\cup{\color{red}c_5}$.
Proceeding similarly for each point leads to the refined bounds  presented in Figure~\ref{fig:twoprofiles}, right. 

\begin{figure}[!hbtp]
	\centering
	\vspace{-3mm}
\begin{tabular}{|c|c|c|}
	\hline
	& \text{\footnotesize{Non-empty intersection}} & \text{\footnotesize{After pruning}} \\ \hline
	${\color{blue}d_1}$ & ${\color{red}c_1,c_2,c_3}$ & ${\color{red}c_2,c_3}$\\ \hline
	${\color{blue}d_2}$ & ${\color{red}c_1,c_2,c_3,c_4,c_5}$ & ${\color{red}c_4,c_5}$\\ \hline
	${\color{blue}d_3}$ & ${\color{red}c_1}$ & ${\color{red}c_1}$ \\ \hline
	${\color{blue}d_4}$ & ${\color{red}c_1,c_2,c_3}$ & ${\color{red}c_2,c_3}$ \\ \hline
	${\color{blue}d_5}$ & ${\color{red}c_4,c_5,c_6}$ & ${\color{red}c_6}$ \\ \hline
	${\color{blue}d_6}$ & ${\color{red}c_4,c_5}$  & ${\color{red}c_4,c_5}$\\ \hline
\end{tabular}

	\vspace{-1mm}
	\caption{Non-empty intersections between confidence intervals, and the resulting pruning.}
	\vspace{-2mm}
	\label{fig:mixprofiles}
\end{figure}

	Even in this simple example, an optimal pruning does not lead to a unique permutation, but four. As the sets $c_i$ and $d_i$ become larger (fewer observations), pruning becomes less effective, keeping more permutations thus yielding less and less refined confidence sets. 
 	Listing all possible permutations to find an optimal pruning by brute force has exponential complexity. Related work on perfect matchings \cite{Fukuda94} can proceed with reduced  computational complexity, yet still linear in the number of the matchings (permutations). Since our goal is not to build an optimal pruning
 	bur rather to use pruning to build refined confidence sets, we now present a computationally less intensive pruning strategy  and then quantify its refinement level.

\section{Building Confidence Sets Exploiting Permutation Equivalence}\label{sec:algo}

In this section, we detail our simple strategy to output confidence sets exploiting $\bG_\cX$-equivalence from an unstructured set of individual confidence sets. It relies on Algorithm~\ref{alg:PrunedMachings} then Algorithm~\ref{alg:RefinedConfidence}.
 
\paragraph{Identification of Compatible Matchings.}

At a high level, the procedure consists in two steps: first building the graph of compatible matchings (non-empty intersections), which can be done in at most $|\cX|^2K^2$ steps.
Second, exploiting the property that permutations are one-to-one to prune the set of compatible matchings (starting from a set $c^k_{i}$ with the smallest number $J$ of matchings $(c^{k'}_{i_j})_{j\leq J}$). This step is in general combinatorial: In the previous example, after removing the obvious assignment of ${\color{blue}d_3}$ to ${\color{red} c_1}$, identifying that the pair
${\color{blue}d_1,d_4}$ should match the pair  ${\color{red}c_2,c_3}$ requires searching for a permutation over a subset of size $2$ (done here using that  ${\color{red}c_2,c_3}$ also matches ${\color{blue}d_4}$). 
More generally, fully exploiting the structure requires searching for permutations over subsets of arbitrary any size, which is computationally demanding (see \cite{Fukuda94}). In order to keep a low computationally, we restrict the size of the permutations the algorithm is looking for to a predefined maximal value $L$ (say $3$), yielding Algorithm~\ref{alg:PrunedMachings}. This results in a pruning whose computational complexity can be controlled.

\vspace{-1mm}
\begin{algorithm}[!hbtp]
	\caption{Prune compatible matchings} \label{alg:PrunedMachings}
	\begin{algorithmic}[1]
		\REQUIRE Confidence sets $(c^k_x)_{x\in\cX,k\in\cK}$, integer $L$.
		\STATE For each $k\!\in\!\cK,x\!\in\!\cX$, for each $k'\!\in\!\cK\!\setminus\!\{k\}$, let
		$\quad\displaystyle{		I_{k,x,k'} = \Big\{ x'\in\cX : c^k_x \cap c^{k'}_{x'} \neq \emptyset \Big\}.}$  \emph{(Matchings)}
		
		\STATE Let $\cJ=\big\{ (k,k',x) : k\in\cK,k'\in\cK,x\in\cX\big\}$  \emph{(Triplets to be examined)}
		\WHILE{$\cJ\neq \emptyset$}
		\STATE Let $(k_0,x_0,k_1)\in\Argmin\big\{|I_{k,x,k'}|\! :  (k,k',x) \!\in\!\cJ\big\}$, \emph{(Pick a candidate with smallest ambiguity)}
		\IF{$|I_{k_0,x_0,k_1}|\leq L$} 
		\STATE Let $\tilde \cX_{x_0} \!=\! \Big\{ x'\!\!\in\!\cX\!:I_{k_0,x',k_1}\!=\!I_{k_0,x_0,k_1}
		 \text{ and } (k_0,k_1,x')\!\in\!\cJ \Big\}$   \emph{(Search for a clique of size $L$)}

		\IF{$|\tilde\cX_{x_0}| = |I_{k_0,x_0,k_1}|$}
		\STATE For each
		$x\!\notin\! \tilde\cX_{x_0}$,  let $\quad\displaystyle{	
		I_{k_0,x,k_1} = \Big\{ x'\in I_{k_0,x,k_1} : x'\notin I_{k_0,x_0,k_1} \Big\}.}$  \emph{(Update the matchings)}
		
		\STATE Set $\cJ = \cJ\setminus \{ (k_0,k_1,x) : x \in \tilde \cX_{x_0}\}$
		\emph{(We are done with this clique)}
		\ELSE
		\STATE $\cJ = \cJ\setminus \{ (k_0,k_1,x_0)\}$ \emph{(We are done with this triplet)}
		\ENDIF
		\ELSE
		\STATE $\cJ = \emptyset$
		\ENDIF
		\ENDWHILE
		\STATE \textbf{Return:} 	$(I_{k,x,k'})_{k,x,k'}$	
	\end{algorithmic}
	\normalsize
\end{algorithm}

\paragraph{Refined Concentration Sets.}
The last step is to build the refined confidence intervals from the set of compatible matchings out put by Algorithm~\ref{alg:PrunedMachings}. Indeed, for each point $x$ and index $k$, it outputs sets $(I_{k,x,k'})_{k'}$, so that each $k'\in \cK\!\setminus\! \{k\}$ may contribute to refining the confidence sets.

\paragraph{Case 1: $I_{k,x,k'}$ is a singleton.}
The situation when $I_{k,x,k'}$ is a singleton, say  $I_{k,x,k'}=\{ x'\}$ is simple to handle, since we then know that $x$ can only be mapped to $x'$; we thus simply denote it $x_{k'}$ in the sequel.
This means we can group together the sample  $\Sam_{k,x}$ (coming from $(x,k)$) and the sample $\Sam_{k',x'}$ (coming from $(x',k')$) to form a novel confidence set. This suggests to introduce
%
\beqan
\cK_{k,x} &=& \Big\{ k'\in\cK\setminus\{k\} : |I_{k,x,k'}|=1 \Big\} \cup \{k\}\, ,
\eeqan
%
and gather all the samples from elements of other distributions that can be uniquely mapped to $x$.
It then remains to compute a novel confidence set using  $\bigcup_{k'\in\cK_{k,x}} \Sam_{k',x_{k'}}$, instead of  $\Sam_{k,x}$, that is to compute $\displaystyle{\CI\bigg(\bigcup_{k'\in\cK_{k,x}} \!\!\Sam_{k',x_{k'}},\delta\bigg)}$.
%
\paragraph{Case 2: $I_{k,x,k'}$ is not a singleton.} When  $|I_{k,x,k'}|>1$, there is an ambiguity to matching $x$ from distribution $k$ to another point related to distribution $k'$.
In the worst case there are $L$ possible matchings for each $k'\neq k$, hence resulting in $L^{K-1}$ possible combinations.
With infinite computational power,  one could form for each combination its corresponding sample, compute the corresponding confidence set combining all the observations in this sample, and then, to account for the ambiguity of the matchings, take the union of all these confidence sets. This would mean to compute $$\displaystyle{\bigcup_{(x_{k'})_{k'}: x_{k'} \in I_{k,x,k'} }\!\!\! \CI\bigg(\bigcup_{k'\in\cK}\Sam_{k',x_{k'}},\delta\bigg)},$$
which is exponential in $K$. In order to avoid this computational blow-up,
we proceed differently: For each $k'\notin\cK_{k,x}$, we do not group observations but build the union of confidence sets $(c_{x'}^{k'})_{x'\in I_{k,x,k'}}$, then simply intersect all the resulting sets. The cost of this computation is $\cO(KL)$ and is no longer exponential in $K$.
Combining the refinements obtained from case 1 and case 2 yields the sets $\overline \CI_{k,x}(\delta)$ summarized in Algorithm~\ref{alg:RefinedConfidence}.

\begin{algorithm}[!hbtp]
	\caption{Refined confidence bounds} \label{alg:RefinedConfidence}
	\begin{algorithmic}[1]
		\REQUIRE Plausible matchings $(I_{k,x,k'})_{k,x,k'}$, confidence level $\delta$.
		\FOR{each $k\in\cK$, $x \in\cX$}
		\STATE Compute 	$\cK_{k,x}= \Big\{ k'\in\cK\setminus\{k\} : |I_{k,x,k'}|=1 \Big\}\cup \{ k\}$
		
		\vspace{-1mm}
		\STATE Compute
		
		\vspace{-6mm}
		$$\quad\overline \CI_{k,x}(\delta) = \CI\bigg(\bigcup_{k'\in\cK_{k,x}} \Sam_{k',x_{k'}},\delta\bigg) \cap \!\!\bigcap_{k'\notin\cK_{k,x}}\!\bigcup_{x'\in I_{k,x,k'}}  \!\!\!c^{k'}_{x'}$$
		
		\vspace{-2mm}
		where $x_{k'}$ denotes the only element of the singleton $I_{k,x,k'}$
		\ENDFOR
		\STATE \textbf{Return:} $(\overline \CI_{k,x}(\delta)  )_{x\in\cX,k\in\cK}$
	\end{algorithmic}
	\normalsize\vspace{-1mm}
\end{algorithm}

\begin{figure*}[!hbtp]
	\centering
	\includegraphics[trim={0mm, 5mm, 0mm, 10mm},clip,width=0.49\textwidth]{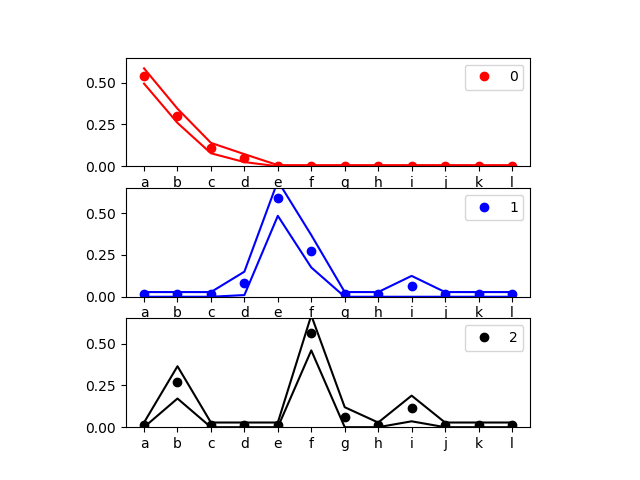}\hspace{-8mm}
	\includegraphics[trim={0mm, 5mm, 0mm, 10mm},clip,width=0.49\textwidth]{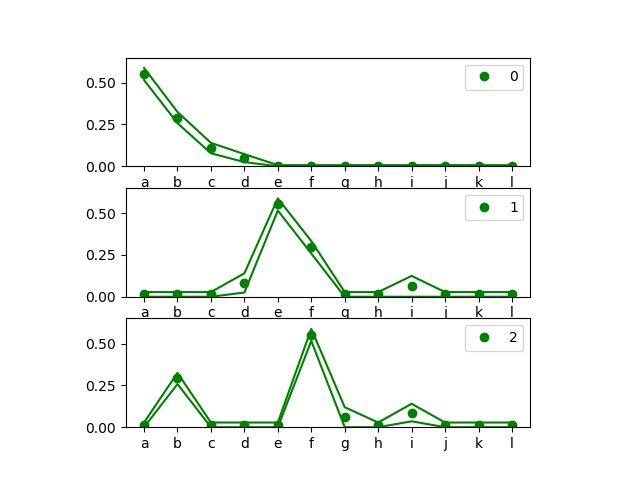}	
	
	\caption{The first experiment. Left: Initial confidence sets, generated from $n_0=1000, n_1=250$, and  $n_2=250$ observations. Right: Confidence sets output by Algorithm~\ref{alg:RefinedConfidence} exploiting $\bG_\cX$-equivalence. 
	}
	\label{fig:algooutput_1}
\end{figure*}

\paragraph{Numerical illustration of Algorithm \ref{alg:RefinedConfidence}.}
We examine the performance of Algorithm \ref{alg:RefinedConfidence} on a small problem with $K=3$ discrete distributions, defined on an alphabet of size $12$, that are $\bG_\cX$-coherent. Our goal is to demonstrate the empirical reduction  in the size of confidence intervals output by Algorithm \ref{alg:RefinedConfidence} over initial ones. In our experiments, we choose $L=5$ for the parameter of Algorithm~\ref{alg:PrunedMachings}. In the first experiment, we choose (see Definition \ref{def:perm_equivalence}) $q = [0.55, 0.3, 0.1, 0.05, 0, 0, 0, 0, 0, 0, 0, 0].$ Figure~\ref{fig:algooutput_1} presents, on the left, the input valid confidence sets, with $\delta=0.1$, obtained respectively from
$n_0=1000$, $n_1=250$, and $n_2=250$  i.i.d.~observations from each distribution and on the right, the
confidence sets (in green) output by Algorithm~\ref{alg:RefinedConfidence}. The dots indicate the true values. As we observe on the last figure, the length of confidence intervals, for the three distributions and for all the elements of $\cX$,
have been significantly reduced. So in this case, using permutation-equivalence allows us to tighten the confidence intervals, thus offering an interesting improvement. We also observe that distributions `1' and `2', which are sampled the least, enjoy the most reduction of their confidence intervals in particular for those elements of $\cX$ whose value of distribution is not close to zero.

\begin{figure*}[!hbtp]
	\centering
	\includegraphics[trim={0mm, 5mm, 0mm, 10mm},clip,width=0.49\textwidth]{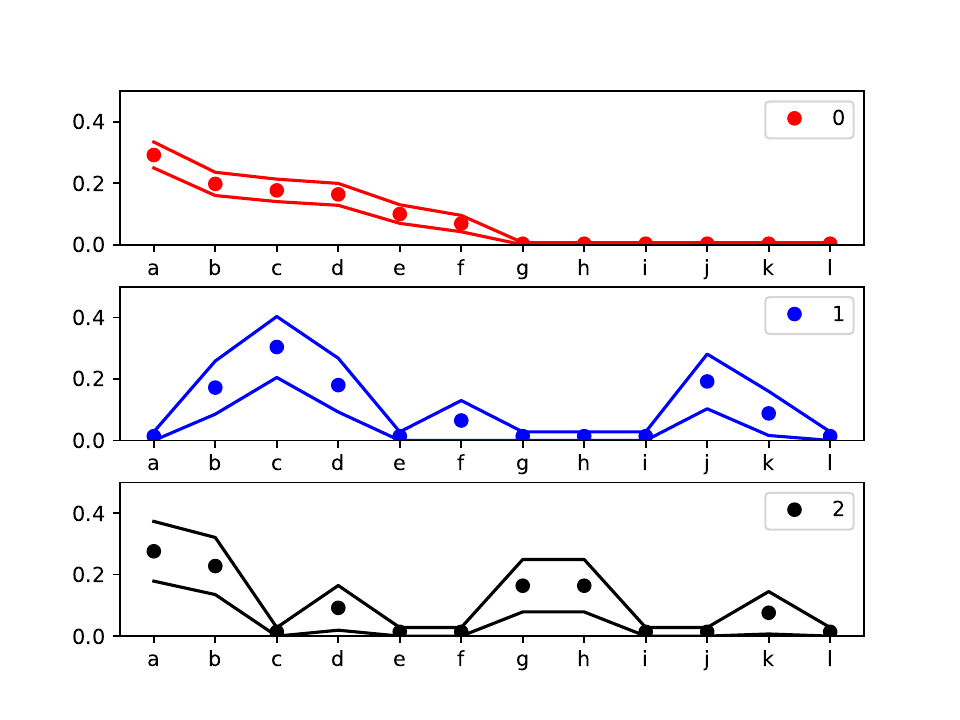}\hspace{-8mm}
	\includegraphics[trim={0mm, 5mm, 0mm, 10mm},clip,width=0.49\textwidth]{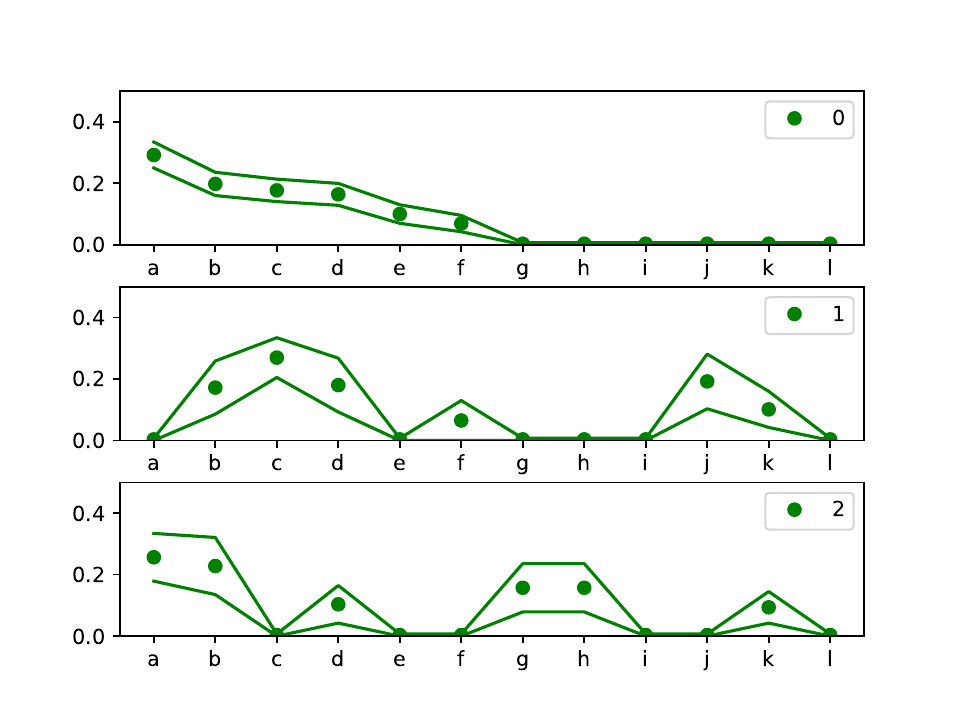}	
	
	\caption{The second experiment. Left: Initial confidence sets, generated from $n_0=1000, n_1=250$, and $n_2=250$ observations. Right: Confidence sets output by Algorithm~\ref{alg:RefinedConfidence} exploiting $\bG_\cX$-equivalence. 
	}
	\label{fig:algooutput_2}
\end{figure*}

In the second experiment, we choose $q = [0.3, 0.2, 0.18, 0.15, 0.1, 0.07, 0, 0,0,0,0,0].$ Similarly to the previous case, Figure~\ref{fig:algooutput_2} presents, on the left, the input confidence sets (with $\delta=0.1$) obtained respectively from
$n_0=1000$, $n_1=250$, and $n_2=250$  i.i.d.~observations from each distribution. It presents, on the right, the
confidence sets output by Algorithm~\ref{alg:RefinedConfidence}.
We note that on this example, where
6 points are outside the support of the distributions, the equivalence enables us to tighten the confidence sets of the points outside the support.
We also note some other tightening, e.g., for point $c$ of distribution $1$, or $k$ of distribution $2$.

%


\section{The Statistical Benefit of Permutation Equivalence}\label{sec:statbene}
The refinement strategy (Algorithms \ref{alg:PrunedMachings}-\ref{alg:RefinedConfidence}) is sound provided that each confidence interval $c^k_{x}=\CI(\Sam_{k,x},\delta)$ is valid, which happens with probability higher than $1\!-\!\delta$, for all $x$ and each $k$.
Let $\Omega:=\{\forall x,\forall k,\, p_k(x) \in  \CI({\Sam}_{k,x},\delta)\}$ be the event that all initial confidence sets are valid. A union bound over all $k\in\cK$ shows that $\Pr(\Omega) \ge 1- K\delta$. The aim of this section is to assess the gain of using refined confidence sets exploiting permutation-equivalence.
In order to simplify the presentation of the results, we consider  $\CI$ given in (\ref{eq:CI_generic}), and introduce the notion of \emph{surrogate confidence intervals}:

\begin{definition}[Surrogate Confidence Intervals]\label{def:surr_CI}
	Let $\mathcal S$ be a sample set, $\delta\in (0,1)$, and consider a confidence interval $\CI$ of the form (\ref{eq:CI_generic}). For some \emph{deterministic} function $B$, we define
	\al{
		\label{eq:CI_canonical_surr}
		\SCI({\Sam},\delta) \!=\!\Big\{ \lambda\! \in\! [0,1]: |\hat p(\Sam) \!-\!\lambda| \!\leq\! B(p,|\Sam|,\delta)\Big\} \,.
	}
		$\SCI$ is a \emph{surrogate confidence interval} (for short, a surrogate) for $\CI$ if $\forall \cS,\delta,\,\CI({\Sam},\delta) \!\subseteq\! \SCI({\Sam},\delta)$.
\end{definition}

For example, $\CI$ and $\SCI$ coincide when $\CI$ is built from Hoeffding inequality. In Section \ref{sec:ex_surr_CI}, we report surrogates of some other standard confidence intervals. The surrogates are only used to simplify the analysis; the algorithm still uses the original confidence sets. 

Our result involves some problem-dependent quantities. For $p\in \cP(\cX)$, let $\cX_{p}\subseteq \cX$. Further, 
for a given set $\cA\!\subseteq\! \cK$, we define $n_\cA \!=\! \sum_{k\in \cA} n_k$ and $\overline{n}_{\cA}\!=\! \max_{k\in \cA} n_k$.
We also make the following mild assumption (it is without loss of generality by density of the set of such distributions):

\begin{assumption}\label{assump:monotone}
	We assume that $q$ is monotone on its support $\cX_{q}$.
\end{assumption}

The following theorems show precisely the interplay between the problem-dependent gaps and the finite number of observations. 
The first one concerns the elements inside the support:

\begin{theorem}[Concentration benefit for $x\in \cX_{p_k}$]\label{thm:permutationequivalencegain}
	Under Assumption \ref{assump:monotone} and the event $\Omega$, it holds for all $k$, for all $x\!\in\! \cX_{p_k}$ (points in the support of $p_k$),
	\beqan
	\overline{\CI}_{k,x}(\delta) \subset \SCI\bigg(\bigcup_{k'\in\tilde \cK_{k,x}} \Sam_{k',x_{k'}},\delta\bigg)\,, \quad \text{ where }
	\eeqan
	%
	\beqan
		\tilde \cK_{k,x}=\{k\} \!\cup\!
		\bigg\{ k'\!\in\!\cK\!\setminus\!\{k\}\!: \forall x'\!\!\in\!\cX\!\setminus\!\!\{x\},  \!\frac{|p_k(x)\!-\!p_{k'}(x')|}{2} \!>\! B\big(p_k(x),n_k,\delta\big) \!+\! B\big(p_k(x'),n_{k'},\delta\big) \bigg\}\,.
	\eeqan
\end{theorem}

We remark that $\tilde \cK_{k,x}$ is a \emph{problem-dependent}, explicit and deterministic set. We now focus on the
remaining points.
%
\begin{theorem}[Concentration benefit for $x\notin \cX_{p_k}$]\label{thm:permutationequivalencegain2}
	Under Assumption \ref{assump:monotone} and the event $\Omega$, it holds for all $k$ and all $x\!\notin\! \cX_{p_k}$,
	\beqan
	\overline \CI_{k,x}(\delta)\!\! &\subset&\!\!\Big\{ \lambda: \lambda \leq B\big(p_k(x),\overline{n}_{\overline \cK_{k}},\delta\big)\Big\}\,,\text{ if } |\cX\setminus\cX_{p_k}|>1\\
	\overline \CI_{k,x}(\delta)\!\! &\subset&\!\!\Big\{ \lambda: \lambda \leq B\big(p_k(x),n_{\overline \cK_{k}},\delta\big)\Big\}\,,\text{ else },
	\eeqan
	with $\overline{\cK}_{k} = \{k\}\cup\big\{ k' \neq k : q_{\min}>q_k(n_{k'})\big\}$, where we introduced  $q_{\min} := \min_{x\in \cX_q} q(x)$ and $q_k(n)= \sup_{x\in\cX_{p_k}}\{2\big(B(p_k(x),n,\delta)+B(0,n_k,\delta)\big) \}$.
\end{theorem}

For a given $k$ and $x$, $\overline\cK_{k}$ and $\tilde\cK_{k,x}$ are explicit deterministic sets (unlike $\cK_{k,x}$) that depend on the unknown distributions and number of observations. They capture the problem-dependent complexity of the problem.
Hence Theorems~\ref{thm:permutationequivalencegain} and \ref{thm:permutationequivalencegain2} guarantee a control of the size of the refined confidence sets
in terms of deterministic problem-dependent sets, that is 
for each instance of the distribution, and each value of each $n_k$. 
This contrasts with purely asymptotic results only showing a speed-up in the limit of a large enough number of observations, as it also enables to capture threshold effects.  
Also, it makes precise the intuition that the improvement increases with the number of observation, and becomes maximal when all initial confidence sets for each distribution are perfectly separated.

\begin{remark}[Asymptotic behavior]
	When $\CI$ is built based on Bernstein's concentration inequality,
	 $b(p,n,\delta)$ scales as $O(\sqrt{p/n})$ for positive $p$, and $O(1/n)$ for $p=0$. Likewise, $q(n)$ scales as $1/n$.  Hence,  we obtain as the corollary the following asymptotic control on the size of the confidence sets
	\beqan
	|\overline \CI_{k,x}| =\begin{cases}
		\widetilde \cO\Big(n_{\tilde \cK_{k,x}}^{-1/2}\Big) & \text{ for }  x\in\cX_{p_k}\,,\\
		\widetilde \cO\Big({\overline{n}^{-1}_{\overline{\cK}_{k}}}\Big) & \text{ for } x\notin\cX_{p_k}\,.
	\end{cases}
	\eeqan
Since for each $x$, the sets $\tilde \cK_{k,x}$ and $\overline{\cK}_{k}$ converge to $\cK$
	as  $\min_k n_k \to \infty$, since entails that the size of the refined confidence intervals is eventually about $\sqrt{\sum_{k'} n_{k'}/n_k}$ smaller for points in the support $\cX_{p_k}$ (which is the best achievable rate),  and $\max_{k'} n_{k'}/n_k$ smaller for points outside the support.
\end{remark}

\begin{figure}[!hbtp]
	\begin{minipage}{0.5\textwidth}
	\includegraphics[trim = {0cm 3mm 15mm 10mm},clip,width=0.99\linewidth]{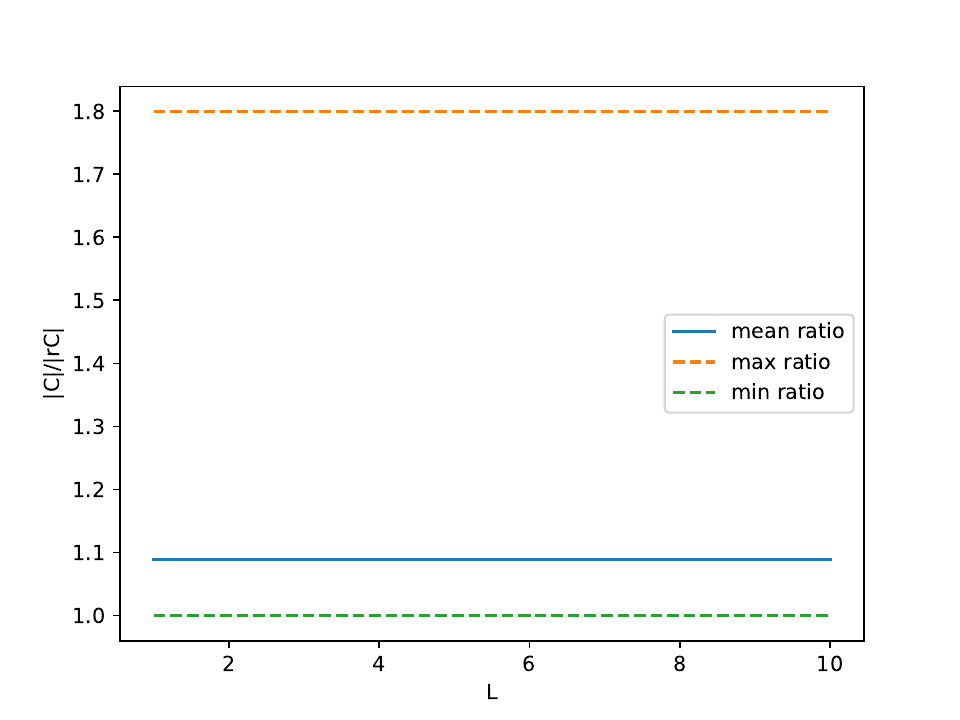}
	\end{minipage}
		\begin{minipage}{0.5\textwidth}	
		\includegraphics[trim={0cm 3mm 15mm 10mm},clip,width=0.99\textwidth]{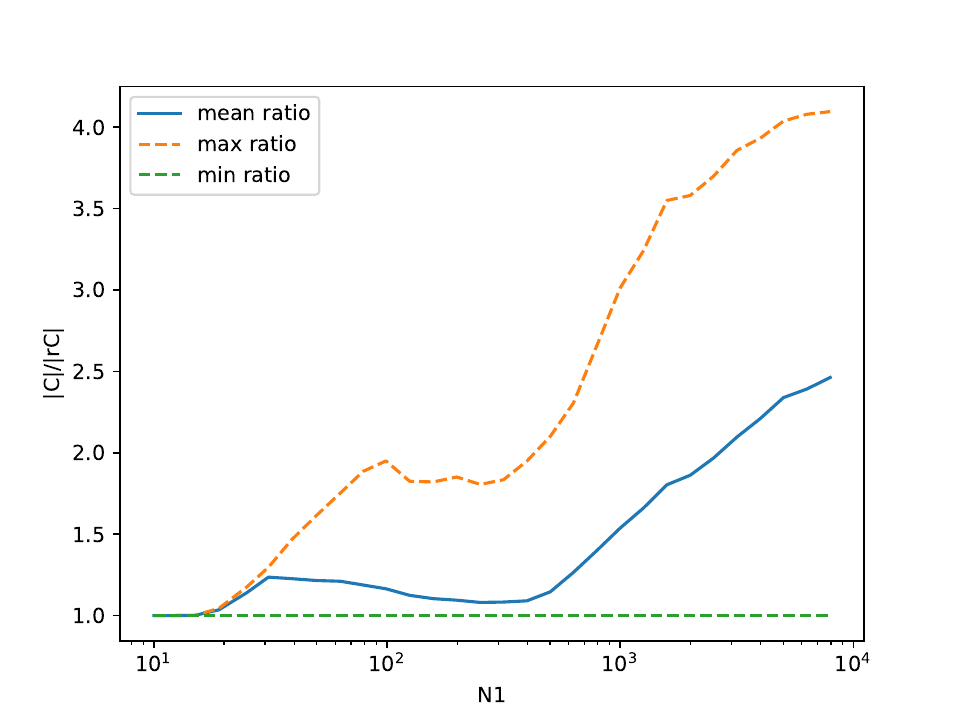}
	\end{minipage}
	\caption{Ratio between initial and refined (empirical Bernstein) confidence sets on problem instances with $|\cX|=10$, $K=5$, 
		as a function of $L$ for $N_1=200$ (left), and as a function of $N_1$ for $L=5$ (right). All values are averaged over $100$ independent experiments.}
	\label{fig:ratios}
\end{figure}

\paragraph{Impact of $L$.}
The acute reader may notice that $L$ does not appear in Theorems~\ref{thm:permutationequivalencegain} and \ref{thm:permutationequivalencegain2}. The reason is that these are worst-case results, stated for Algorithm~\ref{alg:RefinedConfidence} with input plausible matchings not necessarily pruned by Algorithm~\ref{alg:PrunedMachings}. 
Indeed the primary role of $L$ is to reduce the number of tests in the pruning process, hence computations and possibly reduce the set of plausible matchings.
However, this does not imply reduction of confidence sets. Figure~\ref{fig:ratios} actually shows that $L$ little affects the size of the refined confidence sets in practice. Since $L$ significantly
affects computation time of Algorithm~\ref{alg:PrunedMachings}, we suggest the practitioner to scale $L$ to keep it low.

In Figure~\ref{fig:ratios}, we plot the averaged ratio of the initial confidence set
to the refined one over several experiments. We consider an  alphabet $\cX$ of size $10$ and distributions $q$ with support of size $6$. For each $q$, we generate a problem with $K=5$ equivalent distributions, and build the initial and refined confidence sets using empirical Bernstein confidence bounds. One distribution is estimated with $N_1$ observations while all others have $N_1/4$ observations. We then compute for each $x\in\cX$ the ratio between the size of these sets, and store the min, max and average of these values. Finally, we average all results over $100$ randomly generated core distribution $q$. 
Figure~\ref{fig:ratios} (left), shows that considering large values of $L$ (which is computationally demanding), has negligible effect to sharpen the confidence sets. It uses $N_1=200$.
Even for $L=1$, they already divide the size of the initial ones by a factor close to $1.1$ on average and up to $1.8$. Figure~\ref{fig:ratios} (right) shows the effect of $N_1$, for $L=5$.  When $n_k\to\infty$ and under Assumption~\ref{assump:monotone}, confidence sets become uniquely separated  for any value of $L\geq 1$, hence producing maximal shrinkage. When $n_k$ are too small, no improvement is possible. 
The situation in between these situations makes appear a non-trivial behavior, as expected.

There is an intrinsic trade-off between statistical and computational efficiency. For instance here making ``optimal'' use of the structure involves solving an NP-hard problem, which definitely does not scale with $K$ or $X$. 
Algorithm~\ref{alg:RefinedConfidence} aims at solving a relaxation of such NP-hard problem, enjoying a computational complexity of $O(KL)$, as opposed to e.g.~$L^{K}$ or worse complexity when targeting an exact but unnecessary solution. When $O(KL)$ is still considered large, one may further split the set of distributions into smaller groups, with a size growing sublinear in $K$, and still benefit from local speed-up,  gaining computational efficiency at the expense of sacrificing statistical efficiency.

\paragraph{Examples of surrogate sets.}
For completeness and illustration, we present in the following table some confidence intervals constructed using some well-known concentration inequalities, and provide their corresponding surrogate intervals. We detail these derivations in Appendix~\ref{sec:ex_surr_CI}.

\begin{center}
\begin{tabular}{|c|c|c|c|}
	\hline
	Confidence set&$d$ &$b(\lambda,\Sam,\delta)$ in \eqref{eq:CI_generic}&$B(p,|\Sam|,\delta)$ in \eqref{eq:CI_canonical_surr}\\	\hline
	Kullback-Leibler&$\KL$&$f_{\eta}(\delta, |\Sam|)/|\Sam|$&$\sqrt{\!\tfrac{f_{\eta}(\delta, |\Sam|)}{2|\Sam|}}$\\	\hline
	Bernstein&$|\cdot|$&$\sqrt{\frac{2\lambda(1\!-\!\lambda)\log\!\big(\!\tfrac{2|\cX|}{\delta}\!\big)}{|\Sam|}} \!+\! \frac{\log\!\big(\!\tfrac{2|\cX|}{\delta}\!\big)}{3|\Sam|} \,$& $\sqrt{\!\frac{2p(1-p)\log\!\big(\!\tfrac{2|\cX|}{\delta}\!\big)}{|\Sam|}}\! +\! \frac{4.8\log\!\big(\!\tfrac{2|\cX|}{\delta}\!\big)}{|\Sam|}\,$\\	\hline
	Empirical Bernstein&$|\cdot|$&$\sqrt{\!\frac{2\Var(\Sam)\log\!\big(\!\tfrac{4|\cX|}{\delta}\!\big)}{|\Sam|}} \!+\! \frac{7\log\!\big(\!\tfrac{4|\cX|}{\delta}\!\big)}{3|\Sam|} $&$\sqrt{\!\frac{2p(1-p)\log\!\big(\!\tfrac{2|\cX|}{\delta}\!\big)}{|\Sam|}} \!+\! \frac{10\log\!\big(\!\tfrac{2|\cX|}{\delta}\!\big)}{3|\Sam|}$\\	\hline
\end{tabular}
\end{center}
where for $n\in \Nat$, $f_{\eta}(\delta,n) = \eta\log\!\Big(\tfrac{\log(n)\log(\eta n)}{\delta\log^2(\eta)}\Big)$ with $\eta>1$ being an arbitrary peeling parameter.

\begin{figure}[htbp]
	\begin{center}	\includegraphics[width=0.6\linewidth]{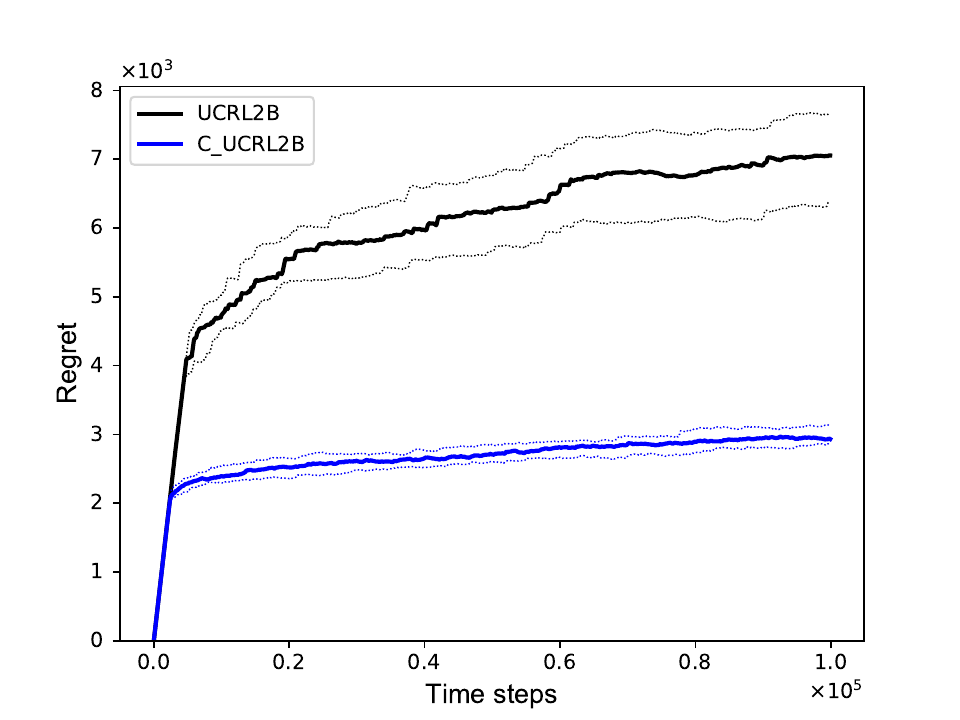}
			\caption{Cumulative regret against a gain-optimal strategy as a function of time for the 6-state River Swim environment, averaged over 50 runs. Comparison between UCRL2B (state-of-the-art), and a version of
			C-UCRL2B from \cite{asadi2019model} using our Algorithm~ \ref{alg:RefinedConfidence} on each equivalent class of distributions.}	
			\label{fig:RLexample}
\end{center}
\end{figure}

\paragraph{An application to Reinforcement Learning.}
We now briefly illustrate the benefit of the proposed algorithm on an undiscounted RL task. We consider the simple and popular RiverSwim environment (see Figure \ref{fig:river_swim} in Appendix), and the average gain optimality criterion \cite{puterman94}, when a learner interacts with an unknown MDP in a single stream of actions and observation until some unknown time horizon.
RiverSwim exhibits a structure of permutation-equivalence
of the next-state transition distributions, as observed already in 
 \cite{asadi2019model}, with $4$ clearly identified classes ($4$ distinct sets of distributions).
In \cite{asadi2019model}, the authors propose a strategy called 
C-UCRL that adapts UCRL2 \cite{jaksch2010near} to incorporate permutation-equivalence, assuming that the underlying permutations $(\sigma_k)_{k\in\cK}$ for each set $\cK$ are \emph{known}. 
However no specific mechanism is proposed to refine the confidence bounds when permutations are unknown. Algorithm~\ref{alg:RefinedConfidence} perfectly applies to this situation. Closely following \cite{asadi2019model}, we easily derive the C-UCRL2B algorithm by adapting UCRL2B \cite{fruit2019improved,jian2019exploration} and making use of our refinement procedure on each of the four group of next-state distributions known to be equivalent, using the same confidence sets as chosen in UCRL2B. We compare this modified strategy against the state-of-the-art UCRL2B in Figure~\ref{fig:RLexample}, showing the substantial reduction of the regret even in this arguably simple environment.

\paragraph{Near-equivalence.}
We have discussed how to exploit the permutation-coherent structure.
In practice, like any structural assumption (Lipschitz smoothness, Linearity, etc.) one may face situations of imperfect structure.
We briefly discuss the case when the equivalence structure is either not exactly known or one can tolerate some error while mistakingly grouping samples from non-equivalent distributions. 
The goal is not to estimate the structure, but rather to provide insights into simple modifications that can help deal with such situations.

Let us consider an 
$\epsilon$-approximation  of the deformation equivalence property, meaning that
only  $\exists \sigma,\forall k, \|p_k-q\circ \sigma_k\|_{TV}\leq \epsilon$ for some known $\epsilon$   is ensured.
Note that if $p_k, p_{k'}$ are not exactly equivalent, one can separate the two distributions asymptotically,
which means that for large number of observations, it happens that $\exists x : I_{k,x,k'}=\emptyset$.  We modify Algorithm~\ref{alg:PrunedMachings}
to set $I_{k,x',k'}=\emptyset$ for all others $x'\!\in\!\cX\!\setminus\!\{x\}$ in that case. When no separation happens, Algorithm~\ref{alg:RefinedConfidence} will produce refined confidence bounds guaranteed to be biased by at most $\epsilon$. In case one can tolerate such a bias, the strategy can be kept unchanged. If one can only tolerate an error of $\eta\!<\!\epsilon$, 
we modify Algorithm~\ref{alg:RefinedConfidence} to first compute $\cK_{k,x}$, then
$$
\cK^\eta_{k,x}\!\!=\! \Big\{ k'\!\!\in\!\!\cK\!\setminus\!\!\{k\}\!\!: |I_{k,x,k'}|\!=\!1 \text{ and } \text{diam}(\CI(\cS_{k', x_{k'}},\delta))\!\leq\! \eta\Big\}\cup \{ k\},
$$
where diam denotes the diameter of the considered interval; we finally redefine the confidence set:
$$\overline \CI_{k,x}(\delta) = \CI\bigg(\bigcup_{k'\in\cK^\eta_{k,x}} \Sam_{k',x_{k'}},\delta\bigg) \cap\! \bigcap_{k'\notin\cK_{k,x}}\bigcup_{x'\in I_{k,x,k'}}  \!\!c^{k'}_{x'}.$$

\section{Conclusion}

In this paper, we have studied the benefit of 
using a \emph{permutation-equivalence} property of a set of unknown distributions $(p_k)_{k\in\cK}$ to  produce a refinement of the confidence sets one may build for each $p_k$ based on observations from $p_k$ only. Leveraging this structure for estimation complements other popular setups involving permutations, such as matching of known distributions (optimal transport), or decision making with structured output. We brought a finite-time analysis
using concentration inequalities to control the potential refinement for each given number of observations $(n_k)_{k\in\cK}$ in a problem-dependent way, and provide an algorithm with low-computational complexity to build the underlying matchings.
Applied to standard Bernstein confidence sets, this enables to get sizes of confidence intervals asymptotically scaling as $O((\sum_k n_k)^{-1/2})$) for points in the support of the distributions and $O((\max_k n_k)^{-1/2})$ for points outside the support, plus to characterize the full finite-time behavior. 
A possible extension of this work is on the one hand to go beyond discrete space $\cX$ and study other automorphisms structures (e.g. rotations for $\cX=\Real^d$, etc.), and on the other hand to apply similar ideas in various machine learning and RL setups.

\bibliography{Bandit_RL_bib,2018library,permutation}
\bibliographystyle{unsrt}


\appendix

\section{The RiverSail Environment}\label{app:riversail}

\begin{figure}[hbtp]
	\begin{center}
		\includegraphics[width=.6\columnwidth]{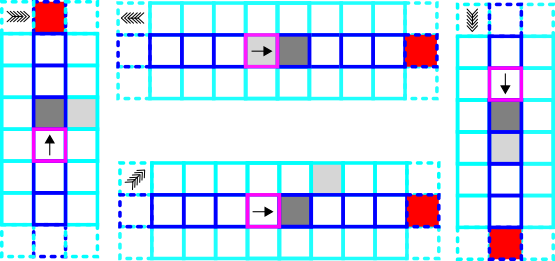}
		\caption{A river-sail environment}
		\label{fig:riversail}
	\end{center}
\end{figure}

We depict in Figure~\ref{fig:riversail} a discrete version of the \emph{RiverSail} environment. It is similar to the windy grid-world \cite{sutton1998reinforcement} or sailboat \cite{epshteyn2008active} environment. In this grid-world MDP, an agent must sail on different rivers while collecting rewards (red states) on the way; exiting a channel (entering dashed states) randomly sends the agent to enter another channel. In each river channel, navigation is similar except for the presence of a constant wind, shown in the top-left corner of each channel, whose direction is unknown, as in a windy windy grid-world: When a boat  in a pink position moves in the direction of the arrow, it ends up in the gray states, shaded according their probability level. All states with dark blue  edges in the same region behave similarly. Here, the exact next-state probability masses are unknown by the sailor, yet she knows perfectly which transitions are equivalent; the permutations between next-state distributions are still unknown due to the unknown wind. Exploiting this structure may massively reduce learning time of the unknown dynamics.

We also provide below an illustration of the RiverSwim environment used in our application to reinforcement learning.
It has been used in \cite{asadi2019model} as well. This is a standard MDP with $L$ states and $2$ actions (left, right).
\begin{figure}[!hbtp]
	\centering
	\footnotesize
	\def\svgwidth{.8\columnwidth}
	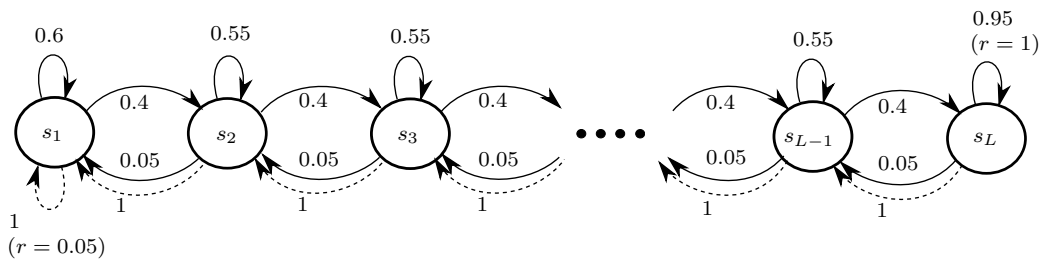
	\caption{The $L$-state \emph{RiverSwim} MDP}
	\label{fig:river_swim}
\end{figure}
The RiverSwim environment has 4 classes of equivalent state-action pairs.
One class for all states and action left, 
one class for states $2$ to $L-1$ with action right,
and one class for each of state $1$ with action right, and state $L$ with action right.
That is, when going right, there is one class for each borders of the river and one for the river itself, and there is one further class for all states when going left.

\section{Benefits of Permutation Equivalence: Proofs}

\begin{myproof}{of Theorem~\ref{thm:permutationequivalencegain} and Theorem~\ref{thm:permutationequivalencegain2}}
	We first handle the points that belong to the support of the considered distributions, then turn to handling the points outside of the support. The reason for doing so is that, by the monotonicity assumption on $q$ (that is, all masses are different), the permutation is uniquely defined on the support. However, it is not uniquely defined outside of the support, which calls for a modified proof.
	
	To simplify the notation, throughout we omit the dependence of various quantities on $\delta$.
	Given $k$ and $x$, denote $\CI_{k,x}:= \CI(\Sam_{k,x},\delta)$, and define $\CI_{k,x}^+ = \max\{\lambda\in \CI_{k,x}\}$ and $\CI_{k,x}^- = \min\{\lambda\in \CI_{k,x}\}$.

	{\bf Step 1: Points  with positive mass.} 
	On the one hand, $(k',x')$ is not compatible with $(k,x)$ if
	\beqan
	\CI_{k,x}^+ < \CI_{k',x'}^- \quad \text{ or }\quad \CI_{k,x}^- > \CI_{k',x'}^+\,.
	\eeqan
	
	In view of the definition of \SCI, this implies that
	\beqan
	\hat p_k(x)+ B( p_{k}(x),n_k) &<& \hat p_{k'}(x')-B(p_{k'}(x'),n_{k'})\\
	\,\,\text{ or }\,\,
	\hat p_k(x)- B( p_{k}(x),n_k) &>& \hat p_{k'}(x')+B(p_{k'}(x'),n_{k'})\,.
	\eeqan
	
	On the other hand,  if $x'$ is compatible with $x$, then it must be with high probability that $p_k(x) + 2B(p_{k}(x),n_k) \geq p_{k'}(x') - 2B(p_{k'}(x'),n_{k'})$, and $p_k(x) - 2B(p_{k}(x),n_k) \leq p_{k'}(x') + 2B(p_{k'}(x'),n_{k'})$,
	or more compactly,
	\beqan
	\frac{|p_{k'}(x')-p_k(x)|}{2} \leq  B(p_{k}(x),n_k)+B(p_{k'}(x'),n_{k'})\,.
	\eeqan

	In particular, since by permutation-coherence it must be that $p_{k'}(x')=p_k(\sigma(x'))$ for some permutation $\sigma$,
	if we assume that
	\beqa
	\forall x_1\in\cX\!\setminus\! \{x\}, \quad \frac{|p_k(x)-p_k(x_1)|}{2} >  B(p_{k}(x),n_k)+ B(p_{k}(x_1),n_{k'}),\label{eqn:condition}
	\eeqa
	then we deduce that whenever $\sigma(x')\in\cX\!\setminus\! \{x\}$, $x'$ cannot be compatible with $x$, so that
	the only point compatible in such a case is $x'$ such that $\sigma(x')=x$.
	By the monotonicity assumption on $q$, then such a point is unique for $x\in\cX_{p_k}$. Note that this also means that $B(p_{k'}(x'),n)=B(p_k(x),n)$.
	
	We deduce that on the event that all confidence bounds are valid,
	\beqan
	\lefteqn{
	\forall x\in\cX_{p_k},\,
	\widetilde \cK_{k,x}\subset \cK_{k,x}, \quad\text{ where } }\\
	&&\!\!\widetilde \cK_{k,x}\!:=\!\{k\}\!\cup\!
	\bigg\{ \!k'\!\in\!\cK\!\setminus\!\{k\}\!:  \forall x_1\!\in\!\cX\!\setminus\! \{x\}, \frac{|p_k(x)\!-\!p_k(x_1)|}{2} \!>\! B(p_{k}(x),n_k)\!+\!B(p_{k}(x_1),n_{k'}) \!\bigg\}\,.
	\eeqan
	
	In particular, since $\CI$ is a decreasing function of the set of samples, it holds that
	\beqan
	\overline \CI_{k,x}\subset \CI\bigg(\bigcup_{k'\in\cK_{k,x}} \Sam_{k',x_{k'}}\bigg)
	\subset  \CI\bigg(\bigcup_{k'\in\widetilde \cK_{k,x}} \Sam_{k',x_{k'}}\bigg)\,.
	\eeqan
	Recalling that $n_{\widetilde \cK_{k,x}} = \sum_{k'\in \widetilde \cK_{k,x}} n_{k'}$, and using the form of $\CI$, we can further write the following
	\beqan
	\forall k,\forall x\in\cX_{p_k},\quad
	\overline \CI_{k,x}\subset \SCI\bigg(\bigcup_{k'\in\widetilde \cK_{k,x}} \Sam_{k',x_{k'}}\bigg) = \bigg\{ \lambda : |\overline p_k(x) - \lambda| \leq B(\lambda, n_{\tilde \cK_{k,x}})\bigg\}\,.
	\eeqan

	{\bf Step 2: Points outside the support.} Let us now deal with points outside of $\cX_{p_k}$. 	
	First, note that when $|\cX\setminus \cX_{p_k}|>1$ and $x\notin\cX_{p_k}$, then there exists another point $x'\notin\cX_{p_k}$.
	Hence for any $k'$, $I_{k,x,k'}$ must contain not only $\sigma(x)$ but also $\sigma(x')$ for some permutation $\sigma$.
	This means that in such a case, $\cK_{k,x} = \{ k \}$. Further we also have $p_k(x)=p_k(x')$.
	Specializing \eqref{eqn:condition} for $x\notin \cX_{p_k}$, it comes that in case
	\beqan
	\forall x_1 \in \cX_{p_k}, \quad \frac{p_k(x_1)}{2}> B(0,n_k)+ B(p_k(x_1), n_{k'})\,,
	\eeqan
	then the only points compatible with $(k,x)$ are such that $x'\notin\cX_{p_{k'}}$, that is $|I_{k,x,k'}| \subset|\cX\setminus \cX_{p_{k'}}|$.
	Hence, let us introduce $q_k(n_{k'})= \sup_{x\in\cX_{p_k}}\{2\big(B(p_k(x),n,\delta)+B(0,n_k,\delta)\big) \}$. We deduce that if for all $x_1 \in \cX_{p_k}, p_k(x_1)>q_k(n_{k'})$, that is $q_{\min}>q_k(n_{k'})$, then
	$|I_{k,x,k'}| \subset|\cX\setminus \cX_{p_{k'}}|$.
	
	Now for each $k'\in\bigg\{ k' \neq k : q_{\min}>q_k(n_{k'})\bigg\}$, the only points compatible with $x\notin\cX_{p_k}$ must be outside of the support of the  distribution $p_{k'}$. In particular, $\hat p_{k'}(x')=0$ for all $x'\in I_{k,x,k'}$.
	This  in turns implies, 	using the definition of the sets $\SCI(\cS) = \{ \lambda : |\hat p(\cS)-\lambda| \leq B(p_k(x),|\cS|)\}$,  that $\bigcup_{x'\in I_{k,x,k'}}  \CI(\Sam_{k',x'})  \subset \bigg\{ \lambda : \lambda \leq B(p_{k'}(x'),n_{k'},\delta )\bigg\}$.
	This motivates to introduce the set
	\beqan
	\overline{\cK}_{k} = \{ k \} \cup \bigg\{ k' \in\cK\setminus\!\!\{k\} : q_{\min}>q_k(n_{k'})\bigg\}\,.
	\eeqan
	If $|\cX\setminus \cX_{p_{k'}}|>1$, then $\cK_{k,x} = \{k\}$ and
	\beqan
	\overline \CI_{k,x} &=& \CI\bigg(\bigcup_{k'\in\cK_{k,x}} \Sam_{k',x_{k'}}\bigg) \cap \bigcap_{k'\notin\cK_{k,x}}\bigcup_{x'\in I_{k,x,k'}}  \CI(\Sam_{k',x'})\\
	&\subset& \CI(\Sam_{k,x}) \cap \bigcap_{k'\neq k : q_{\min}>q(n_{k'})}\bigcup_{x'\in I_{k,x,k'}}  \CI(\Sam_{k',x'})\\
	&\subset& \bigg\{ \lambda: \lambda \leq B(p_k(x),n_{k})\bigg\}\cap \bigcap_{k'\neq k : q_{\min}>q(n_{k'})} \bigg\{ \lambda: \lambda \leq B(p_{k'}(x),n_{k'})\bigg\}\\
	&=&  \bigcap_{k'\in \overline{\cK}_{k}} \bigg\{ \lambda: \lambda \leq B(p_{k'}(x),n_{k'})\bigg\}\\
	& = &\bigg\{ \lambda : \lambda \leq B(p_{k}(x),\overline{n}_{\overline \cK_{k}})\bigg\},
	\eeqan
	where we used the fact that $B(p,\cdot)$ is a non-increasing function.
	Now if  $|\cX\setminus \cX_{p_{k'}}|=1$, then we deduce that $\widetilde \cK_{k,x}=\{k\}\cup	\big\{ k'\in\cK\!\setminus\!\{k\}:  q_{\min}>q_k(n_{k'})\big\} =
	\overline{\cK}_{k} $.
	\beqan
	\overline \CI_{k,x} &\subset& \CI\bigg(\bigcup_{k'\in\tilde \cK_{k,x}} \Sam_{k',x_{k'}}\bigg)\subset \bigg\{ \lambda : \lambda \leq B(p_k(x), n_{\overline{\cK}_{k}})\bigg\}\,.
	\eeqan
	
\end{myproof}

\section{Examples of Surrogate Sets}\label{sec:ex_surr_CI}
In this section, we briefly mention some confidence intervals constructed using some well-known concentration inequalities, and derive their corresponding surrogate intervals, summarized in the table at the end of Section~\ref{sec:statbene}.

\paragraph{Kullback-Leibler (KL) confidence sets.} Using the concentration inequality presented in \cite{MaillardHDR} (see also \cite{garivier2011klucb} and \cite{cappe2013kullback}) for the control of KL deviations for Bernoulli random variables, one can define the following confidence set:
\als{
	\CI(\Sam,\delta) = \Big\{ \lambda \in [0,1]: \kl(\hat p(\Sam),\lambda) \leq \alpha_{\eta}(|\Sam|,\delta)/|\Sam|\Big\}\,,
}

\vspace{-3mm}\noindent
where for $n\in \Nat$, 
$\alpha_{\eta}(|\Sam|,\delta) := \eta\log\Big(\tfrac{\log(n)\log(\eta n)}{\delta\log^2(\eta)}\Big)$
with $\eta>1$ being an arbitrary peeling parameter, and where $\kl(u,v)$ denotes the KL divergence between two Bernoulli distributions with parameters $u$ and $v$:
$
\kl(u,v)=u\log(u/v)+(1-u)\log((1-u)/(1-v)).
$
The confidence sets above admits the generic form (\ref{eq:CI_generic}) with $d$ being the KL divergence $\kl$.  
Using Pinsker's inequality $\kl(x,y)\geq 2(x-y)^2$ valid for all $x,y\geq 0$ gives the following surrogate:
\als{
	\SCI(\Sam,\delta) = \Big\{\lambda \in [0,1]: |\hat p(\Sam) - \lambda| \leq \sqrt{\tfrac{\alpha_{\eta}(|\Sam|,\delta)}{2|\Sam|}}\Big\}\, .
}

\paragraph{Bernstein confidence sets.}The Bernstein concentration inequality for bounded random variables in $[0,1]$ directly leads to a confidence set taking the the generic form (\ref{eq:CI_generic}) with $b=b^{\texttt{Berns}}$,
\als{
	b^{\texttt{Berns}}(\lambda,\Sam,\delta) = \sqrt{\frac{2\lambda(1\!-\!\lambda)\log\big(\tfrac{2|\cX|}{\delta}\big)}{|\Sam|}} + \frac{\log\big(\tfrac{2|\cX|}{\delta}\big)}{3|\Sam|} \, .
}
The following lemma presents a sharp $\SCI$ for such a set: 

\begin{lemma}
	\label{lem:Bernstein_surrogate}
	Consider the Bernstein confidence set described above. Then, $\SCI$ in (\ref{eq:CI_canonical_surr}) with 
	\als{
		B(p,|\Sam|,\delta) = \sqrt{\frac{2p(1-p)\log\big(\tfrac{2|\cX|}{\delta}\big)}{|\Sam|}} + \frac{4.8\log\big(\tfrac{2|\cX|}{\delta}\big)}{|\Sam|}\, ,
	}
	constitutes a corresponding $\SCI$.  
\end{lemma}

\paragraph{Empirical Bernstein confidence sets.}Similarly to the previous case, we can use the empirical Bernstein concentration presented in \cite{maurer2009empirical} to define a confidence set. This set takes the generic form (\ref{eq:CI_generic}) with $b=b^{\texttt{emp-Berns}}$ defined as
\als{
	b^{\texttt{emp-Berns}}(\lambda,\Sam,\delta) = \sqrt{\frac{2\Var(\Sam)\log\big(\tfrac{4|\cX|}{\delta}\big)}{|\Sam|}} + \frac{7\log\big(\tfrac{4|\cX|}{\delta}\big)}{3|\Sam|} \, ,
}
where $\Var(\Sam)$ denotes the empirical variance of $p$ built using $\Sam$. Then: 

\begin{lemma}
	\label{lem:emp_Bernstein_surrogate}
	For the empirical Bernstein confidence set described above. Then, $\SCI$ in (\ref{eq:CI_canonical_surr}) with 
	\als{
		B(p,|\Sam|,\delta) = \sqrt{\frac{2p(1-p)\log\big(\tfrac{2|\cX|}{\delta}\big)}{|\Sam|}} + \frac{10\log\big(\tfrac{2|\cX|}{\delta}\big)}{3|\Sam|}\, ,
	}
	constitutes a corresponding $\SCI$.  
\end{lemma}

In this section, for the sake of completeness and practical guidance, we discuss a few standard concentration results valid for $n$ i.i.d. observations. We then provide some perhaps less known concentration results that are valid uniformly over all number $n$ of observations. They all help build the initial confidence sets $\CI(\Sam,\delta)$.

\begin{myproof}{of Lemma \ref{lem:Bernstein_surrogate}}
	
	For brevity, let us introduce $\zeta := \tfrac{1}{|\cS|}\log\big(\tfrac{2|\cX|}{\delta}\big)$. To prove the lemma, we first show that if $u$ and $\lambda$ satisfy $|u - \lambda| \leq \sqrt{2\lambda(1-\lambda)\zeta} + \zeta/3$, then
	\als{
		(i)& \quad 
		\sqrt{\lambda(1 - \lambda)} \leq \sqrt{u(1- u)} + 2.4\sqrt{\zeta} \, , \\
		(ii)& \quad 
		\sqrt{u(1- u)} \leq  \sqrt{\lambda(1 - \lambda)} + \sqrt{\tfrac{1}{2}\zeta} \, .
	}

	Now if $\CI$ holds, then (i) implies that  
	\als{
		|\widehat{p}(\cS) - \lambda| &\leq \sqrt{2\zeta}\Big(\sqrt{\widehat{p}(\cS)\big(1- \widehat{p}(\cS)\big)} + 2.4\sqrt{\zeta}\Big) + \zeta/3\\
		&\leq \sqrt{2\zeta\widehat{p}(\cS)\big(1- \widehat{p}(\cS)\big)} + 3.8\zeta\, .
	}
	Moreover when $\CI$ holds, 
	$|\widehat{p}(\cS) - p| \leq \sqrt{2p(1-p)\zeta} + \zeta/3$, so that using (ii), we get  $\sqrt{\widehat{p}(\cS)\big(1- \widehat{p}(\cS)\big)} \leq \sqrt{p(1-p)} + \sqrt{\tfrac{1}{2}\zeta}$. Putting together and using some calculations, on the event that $\CI$ holds, we get the desired result:
	\als{
		|\widehat{p}(\cS) - \lambda| \leq \sqrt{2p(1-p)\zeta} + 4.8\zeta\, .
	}

	Now we turn to proving (i) and (ii). 
	
	\paragraph{Proof of (i).}By Taylor's expansion, we have
	\als{
		\lambda(1-\lambda) &= u(1- u) + (1 - 2u)(\lambda - u) -(\lambda- u)^2 \\
		&= u(1- u) + (1 - u - \lambda)(\lambda - u) \\
		&\leq
		u (1-u) + |1 - u - \lambda |\left(\sqrt{2\lambda (1-\lambda)\zeta} + \tfrac{1}{3}\zeta\right) \sk
		&\leq
		u (1-u) + \sqrt{2 \lambda(1- \lambda)\zeta} + \tfrac{1}{3}\zeta
		\, .
	}
	Using the fact that $a\le b\sqrt{a}+c$ implies $a\le b^2 + b\sqrt{c} + c$ for nonnegative numbers $a,b$, and $c$, we get
	\begin{align*}
	\lambda(1-\lambda)
	&\leq
	u (1-u)  + \tfrac{1}{3}\zeta + \sqrt{2\zeta\left(u (1-u)  + \tfrac{1}{3}\zeta\right)} + 2\zeta \sk
	&\leq
	u (1-u)  + \sqrt{2\zeta u (1-u)} + 3.15\zeta  \sk
	&= \left(\sqrt{u (1-u)}  + \sqrt{\tfrac{1}{2}\zeta}\right)^2 + 2.65\zeta\, ,
	\end{align*}
	where we have used $\sqrt{a+b}\le \sqrt{a} + \sqrt{b}$ valid for all $a,b\ge 0$. Taking square-root from both sides and using the latter inequality give the desired result:
	\begin{align*}
	\sqrt{\lambda(1 - \lambda)} &\leq
	\sqrt{u(1- u)} + \sqrt{\tfrac{1}{2}\zeta} + 2.65\sqrt{\zeta} \leq \sqrt{u(1- u)} + 2.4\sqrt{\zeta} \, .
	\end{align*}
	
	\paragraph{Proof of (ii).}Similarly to the previous case, by Taylor's expansion, we have
	\als{
		u(1- u) &= \lambda(1-\lambda) + (1 - 2\lambda)(u - \lambda) -(\lambda- u)^2 \\
		&= \lambda(1-\lambda) + (1 - u - \lambda)(u - \lambda) \\
		&\leq
		\lambda(1-\lambda) + |1 - u - \lambda |\left(\sqrt{2\lambda (1-\lambda)\zeta} + \tfrac{1}{3}\zeta\right) \sk
		&\leq
		\lambda(1-\lambda) + \sqrt{2 \lambda(1- \lambda)\zeta} + \tfrac{1}{3}\zeta \\
		&\leq \Big(\sqrt{\lambda(1 - \lambda)} + \sqrt{\tfrac{1}{2}\zeta}\Big)^2
		\, ,
	}
	which after taking square-root from both sides gives the desired result.   
\end{myproof}

\begin{myproof}{of Lemma \ref{lem:emp_Bernstein_surrogate}}
	
	By the Bernstein inequality, we have $
	|\widehat{p}(\Sam) - p|\leq \sqrt{2p(1-p)\zeta} + \zeta/3$, with probability at least $1-\delta$, 
	where $\zeta := \tfrac{1}{|\cS|}\log\big(\tfrac{2|\cX|}{\delta}\big)$. Hence using inequality (ii) in the proof of Lemma \ref{lem:Bernstein_surrogate}, we have:
	$$
	\sqrt{\widehat{p}(\cS)\big(1- \widehat{p}(\cS)\big)} \leq \sqrt{p(1-p)} + \sqrt{\tfrac{1}{2}\zeta}, \quad \text{with probability at least $1-\delta$.}
	$$
	We therefore deduce that when $\CI$ holds, with probability at least $1-\delta$, 
	\als{
		|\widehat{p}(\Sam) - \lambda|\leq \sqrt{2p(1-p)\zeta} + \tfrac{10}{3}\zeta\, .
	}
	
\end{myproof}

\section{Examples of Confidence Sets}\label{app:confexamples}
In this section, for the sake of completeness and practical guidance, we discuss a few standard concentration results valid for $n$ i.i.d. observations. We then provide some perhaps less known concentration results that are valid uniformly over all number $n$ of observations. They all help build the initial confidence sets $\CI(\Sam,\delta)$.

\subsection{A Few Classical Concentration Inequalities}
\paragraph{Sub-Gaussian confidence sets.}
We first recall that if $(X_i)_{i\leq n}$ are i.i.d.~according to a distribution $\nu$ with mean $\mu$, that is $\sigma$-sub-Gaussian, meaning

\beqan
\forall \lambda\in\Real,\quad \log\Esp\exp(\lambda(X-\mu)) \leq \frac{\lambda^2\sigma^2}{2}\,,
\eeqan
then it holds by the Chernoff-method that

\beqan
\forall \delta\in(0,1),\quad
\Pr\bigg(\frac{1}{n}\sum_{i=1}^n X_i - \mu \geq \sqrt{\frac{2\sigma^2\log(1/\delta)}{n}}\bigg) \leq \delta\,.
\eeqan

\begin{remark}
	Let us recall that distributions with bounded observations in [0,1] are $1/2$-sub-Gaussian.
\end{remark}

\paragraph{Bernstein confidence sets.}
Considering that  $(X_i)_{i\leq n}$  are i.i.d. bounded in [0,1] with variance $\sigma^2$, then a Bernstein inequality yields, for all $\delta\in [0,1]$,
\als{
	\Pr\bigg(\frac{1}{n}\sum_{i=1}^nX_i - \mu \geq \sqrt{\frac{2\sigma^2}{n}\log\big(\tfrac{1}{\delta}\big)} + \frac{\log\big(\tfrac{1}{\delta}\big)}{3n}\bigg) \leq \delta.
}

\paragraph{Bernoulli confidence sets.} 
Since Bernoulli random variables with mean parameter $\mu\in[0,1]$ are  bounded in $[0,1]$ with variance $\mu(1-\mu)$,
sub-Gaussian bounds apply with $\sigma=1/2$, and Bernstein bounds apply with $\sigma^2 = \mu(1-\mu)$.
On the other hand, it holds by a direct application of Cram\'er-Chernoff method that for all $\epsilon\geq0$,

\als{
	\Pr\bigg(\frac{1}{n}\sum_{i=1}^n X_i - \mu \geq \epsilon\bigg) \leq \exp\Big(-n\kl(\mu+\epsilon,\mu)\Big)\,,
}
where $\kl(u,v)$ denotes the Kullback-Leibler divergence between two Bernoulli distributions with parameters $u$ and $v$:
$
\kl(u,v)=u\log(u/v)+(1-u)\log((1-u)/(1-v)).
$
The reverse map of the Cram\'er transform $\epsilon\mapsto \kl(\mu+\epsilon,\mu)$ is unfortunately not explicit, and one may consider its Taylor's approximation to derive approximate but explicit high-probability confidence bounds; see \cite{berend2013concentration,kearns1998large,raginsky2013concentration}. More precisely, it is possible to derive the following sub-Gaussian control of the tails of Bernoulli observations:
\begin{lemma}
	\textnormal{\bf (Sub-Gaussianity of Bernoulli random variables, see e.g.~\cite{berend2013concentration})}
	For all  $\mu\in[0,1]$, the left and right tails of the Bernoulli distribution are controlled in the following way
	
	\beqan
	\forall \lambda \in\Real, &&
	\log \Esp_{X\sim\cB(\mu)} \exp(\lambda (X-\mu)) \leq \frac{\lambda^2}{2} g(\mu)\,,
	\eeqan
	where $g(\mu) = \frac{1/2-\mu}{\log(1/\mu-1)}$.
	The control on right-tail can be further refined when $\mu\in[\frac{1}{2},1]$, as follows:
	
	\als{
		\forall \lambda \in\Real^+,\,\,
		\log \Esp_{X\sim\cB(\mu)}  \exp(\lambda (X-\mu)) \leq \frac{\lambda^2}{2} \mu(1-\mu)\,.
	}
\end{lemma}

As an immediate corollary, introducing the function $\underline{g}(\mu) =\begin{cases} g(\mu)&\text{if }\mu<1/2\\ \mu(1-\mu) &\text{else}\end{cases}$, we obtain

\als{
	\forall \delta\in(0,1),\,\,\,
	\Pr\bigg(\frac{1}{n}\sum_{i=1}^n X_i - \mu \geq \sqrt{\frac{2\underline{g}(\mu)\log(1/\delta)}{n}}\bigg) \leq \delta\,,\\
	\forall \delta\in(0,1),\,\,\,
	\Pr\bigg(\mu- \frac{1}{n}\sum_{i=1}^n X_i \geq \sqrt{\frac{2g(\mu)\log(1/\delta)}{n}}\bigg) \leq \delta\,.
}
These inequalities yield  $\CI(\Sam,\delta)$ from Section~\ref{sec:setup} using

\als{
	&b^{\texttt{sub-G}}(\mu,|\Sam|,\delta)= \sqrt{\frac{2\sigma^2}{|\Sam|}\log\big(\tfrac{2|\cX|}{\delta}\big)}\,,\\
	&b^{\texttt{Bern}}(\mu,|\Sam|,\delta)= \sqrt{\frac{2\mu(1\!-\!\mu)}{|\Sam|}\log\big(\tfrac{2|\cX|}{\delta}\big)} + \frac{\log\big(\tfrac{2|\cX|}{\delta}\big)}{3|\Sam|} \,,\\
	&b^{\texttt{sub-G-Bern}}(\mu,|\Sam|,\delta)= \sqrt{\frac{2g(\mu)}{|\Sam|}\log\big(\tfrac{2|\cX|}{\delta}\big)}\,.
}

\subsection{Time-uniform Confidence Sets}
In a number of machine learning applications, such as when sampling of observations is done actively (e.g.~active learning, multi-armed bandits, reinforcement learning), it is often desirable to obtain concentration bounds
that are not only valid with high probability for each $n$, but rather with high probability over all $n\in\Nat$ simultaneously.  In order to build time-uniform concentration inequalities, one may resort to two main tools (see, e.g., \cite{MaillardHDR}).
In the Gaussian setup for instance, one may resort to the mixture method from \cite{pena2008self},
while in general, a time-peeling proof technique can be considered. While
the time-peeling technique leads asymptotically better bounds, the method of mixture yields usually tighter bounds for small to moderate values of $n$.

\paragraph{Time-uniform sub-Gaussian confidence sets.}
For $\sigma$-sub-Gaussian observations, it holds by the mixture method from \cite{pena2008self}, for all $\delta\in(0,1)$,

\beqan
\Pr\bigg(\!\exists n\!\in\!\Nat,\, \Big|\frac{1}{n}\!\sum_{i=1}^n\! X_i \!-\! \mu\Big| \!\geq\! 2\sigma \beta(n,\delta)\bigg) \!\leq \delta\,,
\eeqan
where for $n\in \Nat$, $\beta(n,\delta) :=\sqrt{ \frac{1}{2n}(1+\frac{1}{n})\log(\sqrt{n+1}/\delta)}$.

\paragraph{Time-uniform Bernstein confidence sets.}
Recalling the definition
$$
\alpha_{\eta}(n,\delta) := \eta \log\Big( \frac{\log(n)\log(\eta n)}{\log(\eta)^2} \frac{1}{\delta}\Big),
$$
a time-uniform version of the Bernstein bound yields (see \cite{MaillardHDR}, together with standard approximation of the Cram\'er transform of sub-Gamma distributions \cite{boucheron2013concentration}), for all $\delta\in (0,1)$,

\als{
	\Pr\bigg(\exists n\!\in\!\Nat, \,\frac{1}{n}\!\sum_{i=1}^n \!X_i \!-\! \mu \!\geq\! \sqrt{\frac{2\sigma^2\alpha_{\eta}(n,\delta)}{n}} + \frac{\alpha_{\eta}(n,\delta)}{3n}\bigg) \leq \delta.
}

\paragraph{Time-uniform Bernoulli confidence sets.} 
Finally, adapting the method of mixtures to the fact that the Bernoulli distributions do not have a symmetric sub-Gaussian control, one has


\begin{lemma}\textnormal{\bf (\cite[Corollary 1]{bourel2020tightening})}	\label{lem:subGaussianBernoulliconcentration}
	Let $(X_i)_{i\leq n}\stackrel{\text{i.i.d.}}{\sim}\cB(\mu)$.
	Then, for all $\delta\in(0,1)$, it holds
	
	\beqan	
	\Pr\bigg(\!\!\exists n\!\in\!\Nat,\, -\!2\sqrt{g(\mu)}\beta(n,\delta)\!\leq\! \frac{1}{n}\!\sum_{i=1}^n \!X_i \!-\! \mu \!\leq\! 2\sqrt{\underline{g}(\mu)}\beta(n,\delta)\!\bigg) \!\leq\! 2\delta\, .
	\eeqan
\end{lemma}

Using these results, one may easily adapt the definition of $\CI(\cS,\delta)$ to the time-uniform setup.

\end{document}

%% file: definitions.tex
\newcommand{\BEAS}{\begin{eqnarray*}}
\newcommand{\EEAS}{\end{eqnarray*}}
\newcommand{\BEA}{\begin{eqnarray}}
\newcommand{\EEA}{\end{eqnarray}}
\newcommand{\BEQ}{\begin{equation}}
\newcommand{\EEQ}{\end{equation}}
\newcommand{\BIT}{\begin{itemize}}
\newcommand{\EIT}{\end{itemize}}
\newcommand{\BNUM}{\begin{enumerate}}
\newcommand{\ENUM}{\end{enumerate}}
\newcommand{\beq}{\begin{equation}}
\newcommand{\eeq}{\end{equation}}
\newcommand{\beqa}{\begin{eqnarray}}
\newcommand{\eeqa}{\end{eqnarray}}
\newcommand{\beqan}{\begin{eqnarray*}}
\newcommand{\eeqan}{\end{eqnarray*}}
\newcommand{\bealn}{\begin{align*}}
\newcommand{\eealn}{\end{align*}}
\newcommand{\al}[1]{ \begin{align} #1  \end{align}}

\newcommand{\als}[1]{ \begin{align*} #1  \end{align*}}

\newcommand{\sk}{\nonumber\\}

\newcommand{\el}{\end{flushleft}}
\newcommand{\bl}{\begin{flushleft}}

\newcommand{\Argmin}{\mathop{\mathrm{Argmin}}}

\newcommand{\bG}{\mathbb{G}}

\newcommand{\cA}{\mathcal{A}}
\newcommand{\cB}{\mathcal{B}}

\newcommand{\cF}{\mathcal{F}}
\newcommand{\cG}{\mathcal{G}}

\newcommand{\cJ}{\mathcal{J}}
\newcommand{\cK}{\mathcal{K}}

\newcommand{\cN}{\mathcal{N}}
\newcommand{\cO}{\mathcal{O}}
\newcommand{\cP}{\mathcal{P}}

\newcommand{\cS}{\mathcal{S}}

\newcommand{\cX}{\mathcal{X}}

\newcommand{\Real}{\mathbb{R}}
\newcommand{\Nat}{\mathbb{N}}

\newcommand{\indic}[1]{\mathbb{I}\{#1\}}

\renewcommand{\phi}{\varphi}
\renewcommand{\epsilon}{\varepsilon}

\newcommand{\Esp}{\mathbb{E}}
\renewcommand{\Pr}{\mathbb{P}}
\newcommand{\Var}{\mathbb{V}}

\newcommand{\kl}{\texttt{KL}}
\newcommand{\KL}{\texttt{KL}}

%% file: jadid_RSE.pdf_tex
\begingroup%
  \makeatletter%
  \providecommand\color[2][]{%
    \errmessage{(Inkscape) Color is used for the text in Inkscape, but the package 'color.sty' is not loaded}%
    \renewcommand\color[2][]{}%
  }%
  \providecommand\transparent[1]{%
    \errmessage{(Inkscape) Transparency is used (non-zero) for the text in Inkscape, but the package 'transparent.sty' is not loaded}%
    \renewcommand\transparent[1]{}%
  }%
  \providecommand\rotatebox[2]{#2}%
  \ifx\svgwidth\undefined%
    \setlength{\unitlength}{604.73983645bp}%
    \ifx\svgscale\undefined%
      \relax%
    \else%
      \setlength{\unitlength}{\unitlength * \real{\svgscale}}%
    \fi%
  \else%
    \setlength{\unitlength}{\svgwidth}%
  \fi%
  \global\let\svgwidth\undefined%
  \global\let\svgscale\undefined%
  \makeatother%
  \begin{picture}(1,0.23520102)%
    \put(0,0){\includegraphics[width=\unitlength,page=1]{jadid_RSE.pdf}}%
    \put(0.93415466,0.12094756){\color[rgb]{0,0,0}\makebox(0,0)[lt]{\begin{minipage}{0.21134427\unitlength}\raggedright $s_L$\end{minipage}}}%
    \put(0.75026781,0.12037184){\color[rgb]{0,0,0}\makebox(0,0)[lt]{\begin{minipage}{0.33040373\unitlength}\raggedright $s_{L-1}$\end{minipage}}}%
    \put(0.93392041,0.23957471){\color[rgb]{0,0,0}\makebox(0,0)[lt]{\begin{minipage}{0.21134427\unitlength}\raggedright $0.95$\\ $(r=1)$\end{minipage}}}%
    \put(0.7594562,0.21904991){\color[rgb]{0,0,0}\makebox(0,0)[lt]{\begin{minipage}{0.21134427\unitlength}\raggedright $0.55$\end{minipage}}}%
    \put(0.84203212,0.15378257){\color[rgb]{0,0,0}\makebox(0,0)[lt]{\begin{minipage}{0.21134427\unitlength}\raggedright $0.4$\end{minipage}}}%
    \put(0.84160615,0.0520501){\color[rgb]{0,0,0}\makebox(0,0)[lt]{\begin{minipage}{0.21134427\unitlength}\raggedright $1$\end{minipage}}}%
    \put(0,0){\includegraphics[width=\unitlength,page=2]{jadid_RSE.pdf}}%
    \put(0.67499125,0.15789925){\color[rgb]{0,0,0}\makebox(0,0)[lt]{\begin{minipage}{0.21134427\unitlength}\raggedright $0.4$\end{minipage}}}%
    \put(0.67494326,0.10360153){\color[rgb]{0,0,0}\makebox(0,0)[lt]{\begin{minipage}{0.21134427\unitlength}\raggedright $0.05$\end{minipage}}}%
    \put(0.67191952,0.05352102){\color[rgb]{0,0,0}\makebox(0,0)[lt]{\begin{minipage}{0.21134427\unitlength}\raggedright $1$\end{minipage}}}%
    \put(0,0){\includegraphics[width=\unitlength,page=3]{jadid_RSE.pdf}}%
    \put(0.84198414,0.09787847){\color[rgb]{0,0,0}\makebox(0,0)[lt]{\begin{minipage}{0.21134427\unitlength}\raggedright $0.05$\end{minipage}}}%
    \put(0,0){\includegraphics[width=\unitlength,page=4]{jadid_RSE.pdf}}%
    \put(0.02976956,0.12452384){\color[rgb]{0,0,0}\makebox(0,0)[lt]{\begin{minipage}{0.33040373\unitlength}\raggedright $s_1$\end{minipage}}}%
    \put(0.02308335,0.22320191){\color[rgb]{0,0,0}\makebox(0,0)[lt]{\begin{minipage}{0.21134427\unitlength}\raggedright $0.6$\end{minipage}}}%
    \put(0.10565928,0.15793457){\color[rgb]{0,0,0}\makebox(0,0)[lt]{\begin{minipage}{0.21134427\unitlength}\raggedright $0.4$\end{minipage}}}%
    \put(0.10561128,0.10099108){\color[rgb]{0,0,0}\makebox(0,0)[lt]{\begin{minipage}{0.21134427\unitlength}\raggedright $0.05$\end{minipage}}}%
    \put(0.10258754,0.0562021){\color[rgb]{0,0,0}\makebox(0,0)[lt]{\begin{minipage}{0.21134427\unitlength}\raggedright $1$\end{minipage}}}%
    \put(0,0){\includegraphics[width=\unitlength,page=5]{jadid_RSE.pdf}}%
    \put(0.19371674,0.22480829){\color[rgb]{0,0,0}\makebox(0,0)[lt]{\begin{minipage}{0.21134427\unitlength}\raggedright $0.55$\end{minipage}}}%
    \put(-0.0012515,0.04109531){\color[rgb]{0,0,0}\makebox(0,0)[lt]{\begin{minipage}{0.21134427\unitlength}\raggedright $1$\\ $(r=0.05)$\end{minipage}}}%
    \put(0.19908608,0.12301893){\color[rgb]{0,0,0}\makebox(0,0)[lt]{\begin{minipage}{0.33040373\unitlength}\raggedright $s_2$\end{minipage}}}%
    \put(0,0){\includegraphics[width=\unitlength,page=6]{jadid_RSE.pdf}}%
    \put(0.2793978,0.15899288){\color[rgb]{0,0,0}\makebox(0,0)[lt]{\begin{minipage}{0.21134427\unitlength}\raggedright $0.4$\end{minipage}}}%
    \put(0.27934981,0.10204939){\color[rgb]{0,0,0}\makebox(0,0)[lt]{\begin{minipage}{0.21134427\unitlength}\raggedright $0.05$\end{minipage}}}%
    \put(0.27897183,0.05726041){\color[rgb]{0,0,0}\makebox(0,0)[lt]{\begin{minipage}{0.21134427\unitlength}\raggedright $1$\end{minipage}}}%
    \put(0,0){\includegraphics[width=\unitlength,page=7]{jadid_RSE.pdf}}%
    \put(0.37520148,0.12309989){\color[rgb]{0,0,0}\makebox(0,0)[lt]{\begin{minipage}{0.33040373\unitlength}\raggedright $s_3$\end{minipage}}}%
    \put(0.36851528,0.22177796){\color[rgb]{0,0,0}\makebox(0,0)[lt]{\begin{minipage}{0.21134427\unitlength}\raggedright $0.55$\end{minipage}}}%
    \put(0,0){\includegraphics[width=\unitlength,page=8]{jadid_RSE.pdf}}%
    \put(0.45393872,0.15912398){\color[rgb]{0,0,0}\makebox(0,0)[lt]{\begin{minipage}{0.21134427\unitlength}\raggedright $0.4$\end{minipage}}}%
    \put(0.45389073,0.10218048){\color[rgb]{0,0,0}\makebox(0,0)[lt]{\begin{minipage}{0.21134427\unitlength}\raggedright $0.05$\end{minipage}}}%
    \put(0.45351275,0.05739151){\color[rgb]{0,0,0}\makebox(0,0)[lt]{\begin{minipage}{0.21134427\unitlength}\raggedright $1$\end{minipage}}}%
    \put(0,0){\includegraphics[width=\unitlength,page=9]{jadid_RSE.pdf}}%
  \end{picture}%
\endgroup%